\documentclass[10pt,journal,compsoc]{IEEEtran}

\usepackage[misc]{ifsym}

\usepackage{mathptmx}
\usepackage[T1]{fontenc}
\usepackage[utf8]{inputenc}
\usepackage[kerning,spacing]{microtype}

\usepackage[numbers,sort&compress]{natbib}
\usepackage{graphicx}
\usepackage[many]{tcolorbox}
\usepackage{mathtools,latexsym,amsfonts,stmaryrd}

\usepackage{booktabs}
\usepackage{pbox}
\usepackage{amsthm}
\usepackage{amsmath}
\usepackage{multirow}
\usepackage{tikz}
\usepackage{mathrsfs}
\usepackage{mathpartir}
\usepackage[ruled,linesnumbered,vlined,noend]{algorithm2e}
\usepackage{url}
\usepackage{syntax}
\usepackage{framed}
\usepackage[noend]{algpseudocode}
\usepackage{listings}
\usepackage{moresize}
\usepackage{wrapfig}
 
\usepackage{seqsplit}
\usepackage{alltt}
\usepackage{xspace}
\usepackage[normalem]{ulem}
\usepackage{enumitem}
\usepackage{pifont}
\newcommand{\cmark}{\ding{51}}

\usepackage[hidelinks]{hyperref}

\DeclareMathAlphabet{\mathcal}{OMS}{cmsy}{m}{n}

\usepackage{thmtools}

\declaretheoremstyle[spaceabove=\topsep,notefont=\normalfont\itshape]{mystyle}

\definecolor{ForestGreen}{RGB}{34,139,34}

\newcommand{\revise}[2]{{\color{red}{\ifx&#1&\else- #1\fi}} {\color{ForestGreen}{\ifx&#2&\else+ #2\fi}}}
\renewcommand{\revise}[2]{#2}

\usetikzlibrary{arrows,patterns, decorations.pathreplacing}

\newcommand{\F}{Fig.}

\newcommand{\T}{Table}
\renewcommand{\S}{Sec.}

\newcommand{\ignore}[1]{}
\newcommand{\mysubref}[2]{\hyperref[#1]{\ref*{#1}(#2)}}

\lstdefinestyle{base}{
  moredelim=**[is][\color{red}]{@}{@},
  escapeinside={<@}{@>}
}

\usepackage{pifont}
\newcommand{\xmark}{\ding{53}}

\newcommand\DejaVuttfamily{%
  \fontfamily{DejaVuSansMono-TLF}\selectfont
}

\lstdefinestyle{base}{
  moredelim=**[is][\color{red}]{@}{@},
  escapeinside={<@}{@>}
}

\lstdefinelanguage
   [x64]{Assembler}
   [x86masm]{Assembler}
   {morekeywords={CDQE,CQO,CMPSQ,CMPXCHG16B,JRCXZ,LODSQ,MOVSXD,%
                  POPFQ,PUSHFQ,SCASQ,STOSQ,IRETQ,RDTSCP,SWAPGS,%
                  rax,rdx,rcx,rbx,rsi,rdi,rsp,rbp,%
                  r8,r8d,r8w,r8b,r9,r9d,r9w,r9b,reg128,m128}}

\usepackage{listings}
\usepackage{fancyvrb}
\usepackage{color}
\definecolor{lightgray}{rgb}{.9,.9,.9}
\definecolor{darkgray}{rgb}{.4,.4,.4}
\definecolor{purple}{rgb}{0.65, 0.12, 0.82}
\definecolor{commentgreen}{RGB}{63,127,95}
\definecolor{pyblue}{RGB}{59,117,175}
\definecolor{pyorange}{RGB}{239,134,54}
\definecolor{pygreen}{RGB}{81,158,62}

\usepackage{xcolor}
\colorlet{myPurple}{blue!40!red}
\definecolor{myOrange}{RGB}{255,192,0}

\lstdefinelanguage{Solidity}{
  keywords={len,delete,int,void,payable, public, event, contract, typeof, new, true, false, catch, function, return, null, catch, switch, var, if, while, do, else, case, break,struct,const,socklen_t,sa_familty_t,char,sockaddr,load},
  keywordstyle=\color{violet}\bfseries,
  ndkeywords={class, export, boolean, throw, implements, import, this},
  ndkeywordstyle=\color{darkgray}\bfseries,
  identifierstyle=\color{black},
  sensitive=false,
  comment=[l]{//},
  escapeinside={(*@}{@*)},
  morecomment=[s]{/*}{*/},
  commentstyle=\color{commentgreen}\ttfamily,
  stringstyle=\color{red}\ttfamily,
  morestring=[b]',
  morestring=[b]"
}

\definecolor{pybg}{RGB}{250,250,250}
\definecolor{pykw}{RGB}{0,112,32}
\definecolor{pystr}{RGB}{186,33,33}
\definecolor{pycmt}{RGB}{96,96,96}
\definecolor{pyself}{RGB}{148,85,141}
\lstdefinestyle{pythontool}{
   language=Python,
   frame=single,
   framerule=0.5pt,
   rulecolor=\color{gray!60},
   backgroundcolor=\color{pybg},
   basicstyle=\scriptsize\ttfamily,
   keywordstyle=\color{pykw}\bfseries,
   stringstyle=\color{pystr},
   commentstyle=\color{pycmt}\itshape,
   showstringspaces=false,
   breaklines=true,
   numbers=none,
   tabsize=4,
   xleftmargin=2pt,
   xrightmargin=2pt,
   aboveskip=2pt,
   belowskip=2pt,
   lineskip=-1pt,
   morekeywords={class,str,BaseTool,dict},
   emph={self,name,description,parameters},
   emphstyle=\color{pyself},
}

\newcommand{\rnum}[1]{\uppercase\expandafter{\romannumeral #1\relax}}

\usepackage{xcolor}

\algnewcommand{\LeftComment}[1]{\Statex \(\triangleright\) #1}

\definecolor{pptbrown}{RGB}{132,60,12}
\definecolor{pptgreen}{RGB}{56,87,35}
\definecolor{pptred}{RGB}{155,30,20}
\definecolor{pptdy}{RGB}{127,96,0}
\definecolor{matplotlib-cyan}{HTML}{17becf}

\newcommand{\parh}[1]{\noindent\textbf{#1}}
\newcommand{\sparh}[1]{\noindent\underline{#1}}  

\newcommand{\tool}{\textsc{agentwm}\xspace}

\newcommand{\rom}[1]{\uppercase\expandafter{\romannumeral #1\relax}}

\theoremstyle{definition}

\title{On Protecting Agentic Systems' Intellectual Property via Watermarking}

\begin{document}

\author{
\IEEEauthorblockN{Liwen Wang$^{1}$, Zongjie Li$^{1}$, Yuchong Xie$^{1}$, Shuai Wang$^{1}$, Dongdong She$^{1}$, Wei Wang$^{1}$, Juergen Rahmel$^{2}$}
\IEEEauthorblockA{$^{1}$The Hong Kong University of Science and Technology\quad $^{2}$HSBC} \\
\IEEEauthorblockA{\{lwanged, zligo, yxiece, shuaiw, dongdong, weiwa\}@cse.ust.hk\quad juergen.rahmel@hsbc.com.hk}
}

\maketitle

\begin{abstract}
 
    The evolution of Large Language Models (LLMs) into agentic systems that
    perform autonomous reasoning and tool use has created significant
    intellectual property (IP) value. We demonstrate that these systems are
    highly vulnerable to imitation attacks, where adversaries steal proprietary
    capabilities by training imitation models on victim outputs. Crucially,
    existing LLM watermarking techniques fail in this domain because real-world
    agentic systems often operate as grey boxes, concealing the internal
    reasoning traces required for verification.
    This paper presents \tool, the first watermarking framework designed
    specifically for agentic models. \tool exploits the semantic equivalence of
    action sequences, injecting watermarks by subtly biasing the distribution of
    functionally identical tool execution paths. This mechanism allows \tool to
    embed verifiable signals directly into the visible action trajectory while
    remaining indistinguishable to users. We develop an automated pipeline to
    generate robust watermark schemes and a rigorous statistical hypothesis
    testing procedure for verification. Extensive evaluations across three
    complex domains demonstrate that \tool achieves high detection
    accuracy with negligible impact on agent performance. Our
    results confirm that \tool effectively protects agentic IP against adaptive
    adversaries, who cannot remove the watermarks without severely degrading the
    stolen model's utility.

\end{abstract}

\section{Introduction}
\label{sec:intro}

\begin{figure}[!t]
    \centering
    \includegraphics[width=0.95\linewidth]{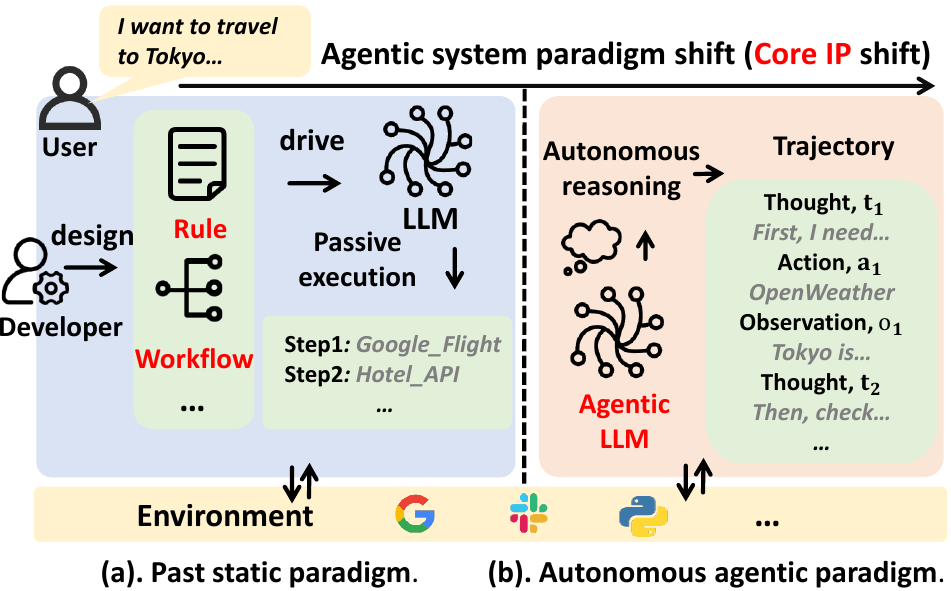}
    \caption{The paradigm shift in agentic system architectures.}
    \label{fig:shift}
\end{figure}

The advent of Large Language Models (LLMs) has transcended the boundaries of
text generation, giving rise to \textit{agentic
systems}~\cite{xi2025rise,wang2024survey,yao2022react}. Unlike traditional
models confined to internal knowledge, these systems use a perception-action
loop to interact directly with external environments, enabling them to achieve
superior performance on complex and reasoning-intensive
tasks~\cite{hendrycksmeasuring,valmeekam2022large}. By orchestrating a diverse
library of tools (e.g., search engines, code interpreters), agentic systems
bridge the gap between text generation and
concrete execution. Given a user query, an agentic system generates a trajectory
of actions (i.e., tool invocations) through continuous interaction with external
environments.

As these agentic systems mature, their underlying architecture has undergone 
a fundamental shift. 
Early agentic frameworks~\cite{patel2024datadreamer,hong2025data,wang2024large} 
relied heavily on developer-crafted workflows with hardcoded execution logic, 
as shown in \F~\ref{fig:shift}(a). In these systems, developers manually 
designed tool orchestration patterns through iterative trial and error, 
while LLMs served merely as passive executors within fixed pipelines~\cite{pan2025multiagent}. 
This paradigm is rapidly evolving with the emergence of reasoning models 
such as OpenAI o3~\cite{openai2025o3} and DeepSeek-R1~\cite{guo2025deepseek}, 
alongside models specifically optimized for tool use, such as MiniMax M2.1~\cite{minimax2025}. 
As illustrated in \F~\ref{fig:shift}(b), modern agentic systems empower LLMs 
to autonomously plan task requirements and dynamically select appropriate tools, 
eliminating the need for predefined workflows. 
This transition fundamentally redefines the nature of Intellectual Property (IP) 
in agentic systems: value has shifted from human-designed orchestration logic 
to the model's intrinsic agentic capabilities. Moreover, developing such 
capabilities requires substantial investment in post-training on high-quality 
datasets~\cite{guo2025deepseek,minimax2025,openai2025o3}, making the agentic 
models themselves valuable proprietary assets that warrant protection.

Prior work demonstrates that LLMs are highly vulnerable to imitation
attacks~\cite{zhao2025survey,liu2025evaluating,carlini2024stealing,li2025differentiation}.
In these attacks, adversaries query a victim model to train a surrogate that
mimics its performance~\cite{carlini2024stealing,li2025differentiation}.
Although agentic systems represent an emerging field, we show that they are
equally susceptible. Extracting agentic capabilities is technically feasible
and proves no more challenging than extracting a standard LLM. As detailed in
\S~\ref{subsec:imitation-attacks-demo}, our empirical analysis confirms that an
extracted imitation agent (denoted as $\mathcal{M}_{imi}$) achieves 95\% of the
victim's performance. These findings expose a critical vulnerability in agentic
systems, highlighting the urgent need to mitigate imitation attacks and prevent
IP theft.

Detecting imitation attacks poses unique challenges. Unlike jailbreak
attacks~\cite{zou2023universal,yi2024jailbreak} or adversarial
inputs~\cite{goodfellow2018making,goodfellow2014explaining} that exhibit
detectable malicious patterns, imitation attacks operate through benign queries
indistinguishable from legitimate user interactions. This makes runtime
detection infeasible without severely degrading user experience through false
positives. Consequently, recent works propose watermarking as a robust
alternative for protecting LLM
IP~\cite{li2023protecting,he2022cater,zhang2024remark,zhao2025can}. These
methods inject WM signals into the victim model's outputs. When an adversary
uses this data to train an imitation model $\mathcal{M}_{imi}$, the WM persists,
serving as verifiable proof of theft. This issue extends beyond academia;
protecting model IP has attracted significant industrial attention. Leading
providers such as Google have deployed watermarking to secure their
proprietary assets~\cite{Dathathri2024}.

However, existing watermarking techniques cannot be directly applied to agentic
systems due to their distinctive operational characteristics. Unlike traditional
LLMs that expose full text outputs, agentic systems generate intermediate steps
(e.g., thought-action-observation) alongside final responses. In
real-world deployments, providers often conceal these intermediate steps to
prevent imitation attacks, exposing only the final actions for billing purposes.
We empirically validate this practice through a survey of 29 commercial agentic
platforms, finding that 82.8\% adopt such grey-box visibility strategies
(Appendix~\ref{appendix:platform-survey}). This
visibility constraint fundamentally undermines existing LLM watermarking
techniques, which depend on access to complete textual outputs for signal
injection and verification. When the majority of tokens are withheld during
inference, watermark (WM) detection becomes infeasible.

This work proposes the first watermarking technique specifically designed for
agentic systems. 
Our approach builds on a key observation: agentic execution
paths contain numerous semantically equivalent action segments. For instance, a
task can be accomplished using either a composite tool or multiple sub-tools in
sequence.
Replacing an action segment with its semantic equivalent yields functionally identical outcomes while preserving high service utility for legitimate users.
Crucially, these mutations are indistinguishable from normal execution patterns,
making them robust against adversarial detection. Based on this insight, we
design \tool, a watermarking framework where each WM corresponds to a
carefully biased distribution $\hat{D}_i$ over equivalent action segments in the
agent's output action trajectory.

We instantiate \tool\ with five watermark schemes in two 
categories: \textit{action-based} schemes exploit equivalence among
individual tool calls (e.g., vendor alternatives, interface aliases), while
\textit{structure-based} schemes leverage equivalent multi-action compositions
(e.g., atomic vs.\ decomposed operations).
Based on these schemes, we develop an automated pipeline
that produces a pool of verified watermark passes from the tool library.
During deployment, each user receives a unique subset of
passes determined by the user ID; activated passes bias matched
action segments from $D_i$ to $\hat{D}_i$ in returned trajectories. For
verification, \tool\ employs statistical hypothesis testing to detect whether a
suspicious model reproduces the watermarked distribution, providing quantifiable
evidence of IP theft.

We evaluate \tool\ across three domains (Social, Data, Business) and diverse
backbone LLMs. Results show that watermarking preserves trajectory quality
($\leq$3\% degradation) and downstream performance (e.g., ARC: 80.09\% vs.\
80.43\%) with only 0.28\% latency overhead. 
\tool\ achieves high
detection (F1 = 1.0) and localizes malicious users with 0.92--1.0 accuracy under
proper configuration.
 Against removal attacks, adversaries achieve F1 $<$ 0.02
in identifying watermark tokens; even fully knowledgeable attackers cannot
remove watermarks without severe quality loss (e.g., $\sim$36\% drop in
trajectory quality). Perplexity~\cite{alon2023detecting} analysis confirms
watermarked outputs are statistically indistinguishable from clean ones.
Ablation studies further validate generalizability across backbone LLMs and
configurations.

\textbf{Contributions.} We summarize our contributions:
\begin{enumerate}[leftmargin=1em, noitemsep]
    \item We formulate the IP protection problem for agentic systems under realistic
    grey-box constraints and present \tool, the first watermarking framework
    designed for this setting.

    \item We propose a distribution-level watermarking approach that exploits
    semantic equivalence in action sequences. We design five complementary
    schemes, an automated pipeline to instantiate watermark passes at scale,
    and a statistical verification procedure for provable IP theft detection.

    \item We evaluate \tool\ across three domains, showing that it preserves
    trajectory quality ($\leq$3\% degradation) with 0.28\% latency overhead,
    achieves high detection (F1 = 1.0), and resists both partially and
    fully knowledgeable adversaries.

\end{enumerate}

\section{Preliminaries and Related Works}
\label{sec:background}

\begin{table}[!tbp]
    \centering
    \caption{Key notation used in the paper.}
    \label{tab:notation}
    \resizebox{0.98\linewidth}{!}{
    \begin{tabular}{c|l}
        \hline
        \textbf{Notation} & \textbf{Description} \\
        \hline
        \multirow{2}{*}{$s_j = (t_j, a_j, o_j)$} & The $j$-th intermediate step,
        including \\
        & thought ($t_j$), action ($a_j$), and observation ($o_j$) \\
        $q_i$; $\mathbf{s}_i = (s_1, \dots, s_n)$ & The $i$-th user query;
        sequence of intermediate steps \\
        $r_i$; $\mathbf{a}_i = (a_1, \dots, a_n)$ & Final response to $q_i$;
        sequence of actions for $q_i$ \\
        $\tau_i = (\mathbf{s}_i, r_i)$; $\tilde{\tau}_i = (\mathbf{a}_i, r_i)$ &
        Complete trajectory; grey-box trajectory \\
        \hline
        $U_{id}$ & A random, unpredictable UID assigned to each user \\
        \multirow{2}{*}{$\mathcal{M}_{vic}$, $\mathcal{M}_{pub}$,
        $\mathcal{M}_{imi}$} & Target agentic model; public backbone LLM; \\ 
         & and attacker's imitation agentic model \\
          \hline
          $D_{\text{ft}}, D_{\text{test}}$ & Fine-tuning dataset split; testing
          dataset split \\
         \hline  
         $P_i, \mathcal{E}_i$ & Each WM pass; the equivalent action segment set in $P_i$
         \\
         $\mathcal{P}_{id}$ & A set of WM passes assigned to user $U_{id}$ \\
         $D_i$ & Distribution of $\mathcal{E}_i$ in $D_{\text{ft}}$ \\
         $\hat{D}_i$ & Tweaked distribution (i.e., WM) of $\mathcal{E}_i$ \\
         \hline  
         $\theta_{J}, \theta_{N}$ & Detection thresholds for JSD and number of
         detected passes \\
         \hline
    \end{tabular}
    }
\end{table}

\subsection{Preliminaries}

\parh{Agentic Systems.}~LLMs' capabilities in natural language understanding and
generation underpin modern agentic
systems~\cite{xi2025rise,wang2024survey,yao2022react}, which tackle complex tasks
through multi-turn interactions with external environments. As shown in
\F~\ref{fig:shift}(a), early implementations relied on developer-crafted
workflows with hardcoded logic, where LLMs served as passive executors within
fixed pipelines, limiting flexibility and generalization.
The emergence of reasoning LLMs (e.g., OpenAI
o3~\cite{openai2025o3}, DeepSeek-R1~\cite{guo2025deepseek}) has shifted this
paradigm (\F~\ref{fig:shift}(b)): instead of following predefined patterns,
modern LLMs autonomously reason and interact with their environment based on
the query and context. Formally, an agentic system processes a user query $q$
to produce a response $r$ through intermediate steps $s_j=(t_j, a_j, o_j)$,
where each step comprises a thought ($t_j$), an action ($a_j$), and an
observation ($o_j$), forming a trajectory $\tau = (\mathbf{s}, r)$ with
$\mathbf{s} = (s_1, \dots, s_n)$.
\T~\ref{tab:notation} summarizes the key notation used throughout this paper.

Building such autonomous capabilities requires intensive training via supervised
fine-tuning (SFT)~\cite{brown2020language,chung2024scaling} and reinforcement
learning (RL)~\cite{ouyang2022training,schulman2017proximal} on curated
datasets~\cite{zhou2023lima,ye2025limo}. This shifts the locus of IP from
explicit, human-defined logic to the internalized reasoning patterns embedded
within agentic models.
To deliver these capabilities, platforms like Coze~\cite{coze2025} and
Dify~\cite{dify2025} offer \textit{domain-specific} agentic services.
However, this creates a transparency-security dilemma: platforms must disclose
tool usage for billing, yet revealing the full trajectory $\tau$ exposes IP to
adversaries~\cite{li2025dissonances,wang2025ip,li2026webcloak}.
To balance both concerns, platforms adopt a \textit{grey-box} visibility setting
(\S~\ref{subsec:threat-model}), revealing only the action sequence $\mathbf{a}$
while concealing internal reasoning $t_j$ and observations $o_j$.
Our survey of 29 commercial agentic platforms (e.g., Coze~\cite{coze2025},
Dify~\cite{dify2025}) confirms that 24
(82.8\%) enforce this grey-box strategy. Details are in
Appendix~\ref{appendix:platform-survey}.

\begin{figure}[!t]
    \centering
    \includegraphics[width=0.9\linewidth]{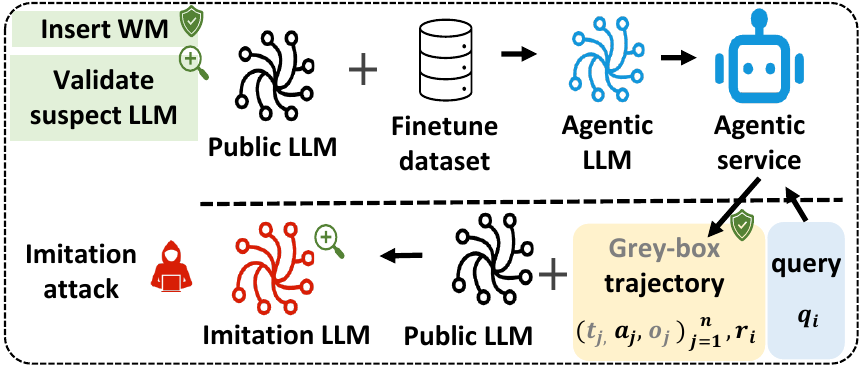}
    \caption{Overview of the imitation attack and mitigation enabled by \tool.}
    \label{fig:background}
\end{figure}

\parh{Trajectory Equivalence.}~A key property of agentic models is
\textit{trajectory equivalence}: given the same query $q$, an agent can produce
different trajectories $\tau_1, \tau_2, \ldots$ that lead to functionally
identical final
responses~\cite{ji2025tree,chen2025optima,zhai2025enhancing,koh2024tree}. 
This stems from the agentic model's inherent decision flexibility: at each step,
multiple valid actions (e.g., different tools or reasoning paths) may satisfy
the same intermediate goal, and these choices cascade into distinct yet
functionally equivalent trajectories~\cite{ji2025tree,chen2025optima,zhai2025enhancing}.
This phenomenon creates a trajectory space for each query $q$, containing
multiple valid execution paths. 
This trajectory diversity enables watermarking: a defender can steer the agentic
model toward specific trajectories as watermark signals, while maintaining
response quality.

\parh{Imitation Attack.}~An imitation attack (also known as model extraction or
model stealing) aims to replicate the behavior of a target model
$\mathcal{M}_{vic}$~\cite{carlini2023extracting,carlini2024stealing,liu2025evaluating,zhao2025survey,li2025differentiation}.
The adversary builds a local imitation model $\mathcal{M}_{imi}$ to bypass
service fees or offer competing services. 
To launch such an attack, the adversary repeatedly interacts with the victim's
domain-specific agentic model. As shown in \F~\ref{fig:background}, the attacker
first prepares a query set $Q$ based on domain-specific knowledge provided by
the platform. For each query $q_i$, the victim model returns a trajectory
$\tau_i$ (or $\tilde{\tau}_i$ under grey-box visibility). 
The attacker then fine-tunes the same public backbone LLM $\mathcal{M}_{pub}$
using the collected dataset $\{(q_i, \tau_i)\}_{i=1}^m$ to produce
$\mathcal{M}_{imi}$. 
Such attacks have proven effective across various domains, including diffusion
models~\cite{carlini2023extracting} and
LLMs~\cite{carlini2024stealing,liu2025evaluating,zhao2025survey,li2025differentiation}.
In many cases, $\mathcal{M}_{imi}$ achieves performance comparable to or even
better than $\mathcal{M}_{vic}$, enabling the adversary to replicate the
victim's intellectual property at minimal
cost~\cite{carlini2023extracting,carlini2024stealing,liu2025evaluating,zhao2025survey,li2025differentiation}.

\subsection{Existing WM Techniques}
\label{subsec:comparison-with-existing-approaches}

Existing LLM watermarking techniques fall into two primary categories.

\parh{Content Watermarking.}~These methods embed watermark signals into token
generation patterns. Early work such as
KGW~\cite{kirchenbauer2023watermark} partitions the vocabulary into ``red'' and
``green'' lists, biasing token selection toward green tokens. Building on KGW,
subsequent efforts have introduced several improvements. For example,
entropy-based sampling~\cite{liu2024adaptive} reduces the impact on text fluency. SIR~\cite{liusemantic}
and Semstamp~\cite{hou2024semstamp} incorporate semantic information into watermark generation,
ensuring that similar semantic content maps to consistent watermark patterns for
improved robustness. Gumbel-Softmax watermarking~\cite{fu2024gumbelsoft} optimizes the biasing
strategy to provably minimize degradation in text quality. 
CoTGuard~\cite{wen2025cotguard} embeds watermark patterns in the model's
CoT reasoning and verifies watermarks by checking consistency
between Chain of Thought (CoT) content and final responses. 
Additionally, agent watermarking remains largely underexplored. The only 
prior work~\cite{huang2025agent} targets a limited setting where 
the LLM outputs action probability values as text, and was evaluated on 
a small action space of six actions in single-step tasks. However, real-world 
agents typically derive actions from logits and handle multi-step tasks, 
limiting the applicability of this approach to practical deployments.

\parh{Model Watermarking.}~These approaches embed watermark 
signals directly into model parameters. One line of work relies on 
white-box verification, where specific parameter characteristics serve as 
identifying fingerprints. For instance, \cite{yoon2025intrinsic} extracts statistical 
properties from model parameters, while EaaW~\cite{shao2024explanation} embeds multi-bit 
watermarks into feature attribution explanations. Another line leverages 
backdoor-based mechanisms, where the model produces predefined responses 
when triggered by specific inputs. For example, Instructional 
Fingerprinting~\cite{xu2024instructional} uses instruction-following tasks to elicit 
deterministic signatures under black-box queries.

\parh{Comparison and Limitations.}~As shown in \T~\ref{tab:comparison},
existing watermarking techniques face significant limitations when applied to
agentic systems.
Text watermarking fails under grey-box constraints: agentic models produce
concise final responses with insufficient tokens for reliable detection, and
watermarking tool calls (e.g., \texttt{code\_interpreter}) corrupts their strict
syntax, breaking functionality.
Model parameter watermarking requires white-box access for verification, which
is impossible when attackers deploy black-box imitation models.
Backdoor-based methods suffer from two critical flaws: backdoor signatures are
easily disrupted during fine-tuning~\cite{xu2024instructional}, and they cannot provide user-level
traceability at scale.
The only prior agent watermarking work~\cite{huang2025agent} addresses a
limited scenario with text-based action probabilities and single-step tasks,
leaving multi-step agentic workflows unaddressed.
These gaps collectively motivate \tool: a watermarking framework designed
specifically for the grey-box, multi-step nature of agentic systems.

\section{Motivation}
\label{sec:motivation}

\subsection{Threat Model}
\label{subsec:threat-model}

\parh{Target Agentic Systems.}~We focus on commercially deployed domain-specific
agentic systems on platforms such as Dify~\cite{dify2025}. As discussed in
\S~\ref{sec:background}, we consider
the grey-box visibility model commonly adopted by commercial agentic platforms.
This model balances IP protection with operational transparency by concealing
internal reasoning steps while disclosing two essential components: (1) the
action sequence $\mathbf{a} = (a_1, \ldots, a_n)$, which documents tool usage
for billing purposes, and (2) the final response $r$, which constitutes the
delivered service output. These components form the grey-box trajectory visible
to users. While some platforms optionally expose partial reasoning or
intermediate tool parameters, such information is often summarized, filtered, or
inconsistently provided, rendering it unsuitable as a reliable detection signal.
Our grey-box model therefore captures the minimal yet reliably available
information across real-world deployments.

Following standard practices in the literature, we assume that neither the
service provider nor potential attackers train foundation models from scratch.
Instead, both parties build upon publicly available backbone LLMs, denoted as
$\mathcal{M}_{pub}$. The service provider fine-tunes $\mathcal{M}_{pub}$ on a
private dataset $D_{\text{ft}}$ containing high-quality domain-specific
trajectory data, producing the victim model $\mathcal{M}_{vic}$ (e.g., a
specialized travel planner or financial advisor).

Notably, we do not assume that the victim and attacker necessarily use identical
backbone models. While prior work on LLM fingerprinting demonstrates that
identifying the backbone model is highly feasible, our framework aims for
broader applicability. In our main experiments, we adopt the conservative
assumption that both parties use the same $\mathcal{M}_{pub}$ to establish a
clear baseline. We then evaluate cross-model generalization in
\S~\ref{subsec:ablation}, where we analyze scenarios involving different backbone
models to demonstrate the robustness of our approach.

\begin{table}[!t]
    \centering
    \caption{Comparison of \tool with existing watermarking techniques.
    \textbf{PF}: Parameter-Free (no access to model parameters required);
    \textbf{GB}: Grey-Box compatible (operates when only action sequences are visible);
    \textbf{AS}: Agent-Supported;
    \textbf{FH}: Functionally Harmless (preserves tool call correctness);
    \textbf{DA}: Domain-Agnostic (generalizes across diverse agentic domains);
    \textbf{UT}: User-level Traceability at scale.
    }
    \label{tab:comparison}
    \resizebox{0.8\linewidth}{!}{
    \begin{tabular}{lcccccc}
    \toprule
    \textbf{Method} & \textbf{PF} & \textbf{GB} & \textbf{AS} & \textbf{FH} & \textbf{DA} & \textbf{UT} \\
    \midrule
    KGW~\cite{kirchenbauer2023watermark} & \cmark & \xmark & \xmark & \xmark & \cmark & \xmark \\
    EaaW~\cite{shao2024explanation} & \xmark & \cmark & \cmark & \cmark & \cmark & \cmark \\
    IF~\cite{xu2024instructional} & \xmark & \cmark & \cmark & \xmark & \cmark & \xmark \\
    \cite{huang2025agent} & \cmark & \cmark & \cmark & \xmark & \xmark & \xmark \\
    \midrule
    \textbf{\tool (Ours)} & \textbf{\cmark} & \textbf{\cmark} & \textbf{\cmark} & \textbf{\cmark} & \textbf{\cmark} & \textbf{\cmark} \\
    \bottomrule
    \end{tabular}
    }
    \end{table}

\parh{Adversary Capabilities and Goals.} 
We assume an adversary capable of querying the target agentic API within a
reasonable budget to facilitate model extraction. As illustrated in
\F~\ref{fig:background}, the adversary's primary objective is to replicate
the proprietary capabilities of the victim model. To achieve this, the attacker
first harvests a collection of grey-box trajectories $\tilde{\tau}$ through
extensive querying. We posit a sophisticated adversary who employs advanced
reconstruction techniques to infer the concealed reasoning steps or extract
intermediate content, leveraging methods such as auxiliary LLM
prompting~\cite{shen2025efficient} or prompt
stealing~\cite{wang2025ip,yang2025prsa}. This assumption represents an
upper-bound threat model, where the adversary successfully reconstructs the
grey-box observations into complete trajectories $\tau$. Subsequently, the
attacker fine-tunes their imitation model $\mathcal{M}_{imi}$ on these
reconstructed trajectories to clone the victim's behavior.

For clarity, we detail the specific assumptions regarding IP theft detection in
\S~\ref{sec:check}. Note that while prior research has investigated the leakage
of private training data from LLMs~\cite{carlini2021extracting}, such work
focuses on privacy violations rather than behavioral cloning. Therefore, those
findings are distinct from and orthogonal to the intellectual property
protection focus of this paper.

\subsection{Imitation Attacks Demo}
\label{subsec:imitation-attacks-demo}

\parh{Setup.}~Imitation attacks on LLMs and diffusion models have become a
well-documented
threat~\cite{carlini2023extracting,carlini2024stealing,liu2025evaluating,zhao2025survey,li2025differentiation}.
To motivate our watermark-based protection, we conduct a proof-of-concept (PoC)
attack on a representative agentic model. We start with
Qwen3-4B~\cite{yang2025qwen3}, a popular open-source LLM, and fine-tune it on
our domain-specific dataset $D_{\text{ft}}$ to create the victim agentic model
$\mathcal{M}_{vic}$. We introduce $D_{\text{ft}}$ and evaluation metrics
in \S~\ref{sec:setup}. For the attack, we query $\mathcal{M}_{vic}$ to collect
grey-box trajectories $\tilde{\tau}$ and reconstruct them into complete
trajectories $\tau$ using an auxiliary LLM, following the standard imitation
attack setup described in \S~\ref{sec:background}. We then fine-tune another
instance of Qwen3-4B on these reconstructed trajectories to obtain the imitation
model $\mathcal{M}_{imi}$.

\begin{table}[!t]
    \centering
    \caption{Performance comparison between victim model, baseline model, and imitation model. PR: Pass Rate, RS: Response Score, TS: Trajectory Score. All metrics are higher is better.}
    \label{tab:attack-results}
    \resizebox{0.7\linewidth}{!}{
    \begin{tabular}{lccc}
    \toprule
    \textbf{Model} & \textbf{PR} & \textbf{RS} & \textbf{TS} \\
    \midrule
    $\mathcal{M}_{vic}$ & 0.687  & 0.574 & 0.903 \\
    \midrule
     $\mathcal{M}_{pub}$ (No imitation) & 0.418 & 0.339 & 0.596 \\
    $\mathcal{M}_{imi}$ (with imitation) & 0.652 & 0.547 & 0.834 \\
    \bottomrule
    \end{tabular}
    }
\end{table}

\parh{Attack Results.}~\T~\ref{tab:attack-results} presents the effectiveness of
the imitation attack. We evaluate model performance using three complementary
metrics (detailed in \S~\ref{sec:setup}). The results demonstrate the threat
posed by imitation attacks. Without imitation, the baseline model
$\mathcal{M}_{pub}$ achieves only about 60\% of the victim's performance.
However, after fine-tuning on stolen trajectories, the imitation model
$\mathcal{M}_{imi}$ achieves performance comparable to that of the victim
model.

Our results in \T~\ref{tab:attack-results} serve as a proof-of-concept to
demonstrate the feasibility of imitation attacks on agentic models. In practice,
real-world attackers often employ advanced
techniques~\cite{chandrasekaran2020exploring,li2025differentiation} to achieve
comparable model performance with significantly fewer queries. This further
amplifies the threat posed by
imitation attacks. While our PoC establishes the viability of such attacks, we
note that not all agentic models are equally vulnerable. Some commercial systems
(e.g., Manus~\cite{manus2025}) completely hide their internal execution details and may
employ additional defenses, making imitation substantially harder. Nevertheless,
our results confirm that imitation attacks pose a realistic and serious threat
to many deployed agentic models, motivating the need for effective watermarking
protection.

\subsection{Design Considerations}
\label{subsec:design-considerations}

Given the unique challenges of watermarking agentic models under grey-box 
constraints, we identify five key properties that an effective watermarking 
solution must satisfy. These considerations guide our design and evaluation 
of \tool.

\parh{Fidelity.}~Watermarking must preserve the agentic model's utility across 
multiple dimensions. First, final response quality must remain unchanged to 
ensure consistent user experience. Second, action sequences must maintain both 
semantic quality and functional equivalence, as any degradation in executable 
logic or API compatibility directly impacts task success rates. Third, the 
model's performance on general domains must be preserved to ensure watermarking 
does not compromise broader capabilities beyond the protected domain.

\parh{Reliability.}~The watermark verification process must achieve high 
true positive rates when detecting imitation models while maintaining low 
false positive rates on unrelated models. This ensures reliable IP verification 
without false accusations. Additionally, the watermark must provide 
traceability to identify specific malicious users who access the victim 
model, enabling effective accountability for IP theft.

\parh{Robustness.}~Watermarks must persist against adversarial removal 
attempts. Attackers may deploy various strategies to eliminate watermark signals 
from stolen trajectories, such as randomly deleting tokens, using LLMs to 
rephrase trajectories, or replacing suspected watermarked elements. The 
watermark should survive these removal attacks without requiring the victim 
model owner to predict specific attack strategies in advance.

\parh{Stealthiness.}~Watermarked trajectories should be indistinguishable 
from normal trajectories to prevent detection and filtering during data 
collection. If attackers can identify watermarked samples, they can discard 
them before training imitation models, rendering the watermark ineffective. 
This requires the watermark to blend naturally with typical behavior patterns.

\parh{Cost.}~Both watermark insertion and verification should incur minimal 
computational overhead. The insertion process should not significantly increase 
inference latency or deployment costs. Verification should be efficient enough 
to enable practical IP enforcement without excessive queries or computational 
resources, ensuring real-world feasibility.

We design \tool\ to satisfy all five considerations. We introduce the technical 
details of \tool\ in the following section and systematically evaluate each 
property in \S~\ref{sec:evaluation}.

\section{Methodology}
\label{sec:methodology}

\begin{figure*}
    \centering
    \resizebox{0.9\linewidth}{!}{
        \includegraphics[width=0.9\linewidth]{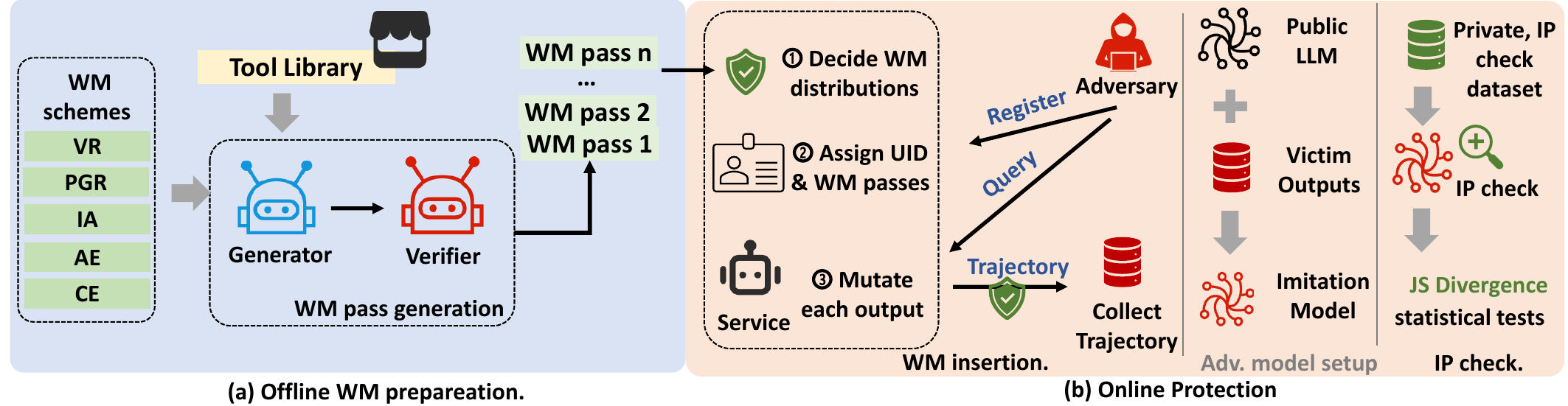}
        } \caption{Overview of \tool. (a) Offline watermark preparation: the
        Generator mines candidate equivalence sets from the tool library with
        five WM schemes, and the Verifier validates them to produce watermark
        passes. (b) Online protection: \tool\ assigns user-specific passes,
        applies watermarks to trajectories, and detects IP theft through
        statistical divergence tests.}
        \label{fig:methodology-overview}
  
\end{figure*}

Our approach is founded on the key observation that multiple equivalent intermediate steps can achieve the same outcome within the agent's trajectory space, as discussed in \S~\ref{sec:background}. 
These sequences may differ syntactically yet preserve functional semantics. We exploit this multi-path property to embed watermarks.
However, the grey-box visibility constraint (\S~\ref{subsec:threat-model}) restricts what users can observe. Embedding watermarks in concealed intermediate reasoning steps would render them unverifiable during IP theft detection. 
We therefore focus exclusively on the action sequence $\mathbf{a}$, which remains consistently visible in the grey-box trajectory $\tilde{\tau} = (\mathbf{a}, r)$.

Specifically, when a potentially malicious user submits a query, the agent first generates a raw trajectory $\tau$. Before returning the grey-box trajectory $\tilde{\tau} = (\mathbf{a}, r)$ to the user, \tool\ intercepts the output and applies a series of watermark passes. The watermark passes transform the original  $\mathbf{a}$ into a watermarked version $\mathbf{a}_{wm}$ by injecting subtle, user-specific patterns while strictly preserving functional semantics. \tool\ then returns $\tilde{\tau}_{wm} = (\mathbf{a}_{wm}, r)$, thereby embedding provenance signals without compromising service utility.

\subsection{WM Definition}
\label{subsec:wm-definition}

We formalize watermarks as induced distribution shifts over the space of
equivalent action sequences.

\parh{Definition 1 (Watermark).}~An action sequence watermark exploits the fact
that equivalent action sequences can produce nearly identical final responses
despite their syntactic differences. 
Formally, a watermark pass $P_i$ is defined by an equivalent action segment set
$\mathcal{E}_i = \{w_1, w_2, \dots, w_k\}$. This set contains distinct action
segments (i.e., sub-sequences of actions) that perform an identical semantic
function. Let $D_i$ denote the natural probability distribution of these
segments in the $D{\text{ft}}$. The watermark $P_i$ transforms an action
sequence $\mathbf{a}$ into $\mathbf{a}'$ by detecting any segment $w \in
\mathcal{E}_i$ and replacing it with a substitute $w' \in \mathcal{E}_i$ sampled
from a biased target distribution $\hat{D}_i$. This replacement embeds a
statistical signature while preserving the functional utility of $\mathbf{a}$.

We design five distinct watermarking schemes that exploit equivalence from
complementary perspectives (\S~\ref{subsec:wm-scheme-taxonomy}). To instantiate
these schemes at scale, we develop an automated pipeline
(\S~\ref{subsec:wm-instance-generation}) to mine equivalence patterns,
generating hundreds of unique watermark passes.

\subsection{WM Insertion Procedure}
\label{subsec:wm-insertion-procedure}

\F~\ref{fig:methodology-overview} illustrates the watermark insertion phase in
\tool. We detail the three key steps below. The verification phase (IP theft
check) is discussed in \S~\ref{sec:check}.

\parh{\ding{172} Deriving Target Distribution $\hat{D}_i$.}~For each watermark
pass $P_i$ corresponding to an equivalence set $\mathcal{E}_i = \{w_1, w_2,
\dots, w_k\}$, we construct a target distribution $\hat{D}_i$. Let $p(w_j)$
denote the natural probability of segment $w_j$ in the victim agent's
original distribution $D_i$. 
Following standard watermarking principles~\cite{li2023protecting}, we apply a logit bias to
shift $D_i$ toward $\hat{D}_i$. Specifically, we strictly designate one segment
$w_t \in \mathcal{E}_i$ as the \textit{target segment} and boost its probability
via exponential scaling. The target probability $\hat{p}(w_j)$ is computed as:
\begin{equation}
    \hat{p}(w_j) = \begin{cases}
    \frac{p(w_t) \cdot e^\delta}{p(w_t) \cdot e^\delta + \sum_{k \neq t} p(w_k)} & \text{if } w_j = w_t \\
    \frac{p(w_j)}{p(w_t) \cdot e^\delta + \sum_{k \neq t} p(w_k)} & \text{otherwise}
    \end{cases}
    \end{equation}
where $\delta$ is a hyperparameter controlling the bias strength. This operation
creates a statistically detectable shift toward $w_t$ while maintaining a valid
probability distribution. We also discuss the impact of $\delta$ in
\S~\ref{subsec:ablation}.

\parh{\ding{173} User Registration and Pass Assignment.}~Upon registration, each
user is assigned a unique $N$-bit User ID (UID) $U_{id}$, where $N$ is the total
number of available watermark passes. We require $U_{id}$ to be random, unique,
and cryptographically unpredictable to attackers.\footnote{This can be
implemented using salted cryptographic hash functions (e.g., SHA-256)~\cite{gilbert2003security}.}
The $U_{id}$ determines the specific subset of watermark passes
$\mathcal{P}_{id}$ activated for that user. The mapping $\mathcal{P}_{id}
\leftarrow U_{id}$ activates pass $P_{i+1}$ if and only if the $i$-th bit of
$U_{id}$ is 1 (for $i \in [0, N-1]$).

Since $\mathcal{P}_{id}$ is unique to each user, it acts as a digital
fingerprint for identifying malicious users involved in IP theft (see
\S~\ref{sec:check}). To balance detection reliability with user capacity, we
constrain the Hamming weight of the UID (i.e., the number of active passes
$|\mathcal{P}_{id}|$) to the range $[5, 20]$. Even in the only one domain
(Data, $N=39$ passes; see \T~\ref{tab:wm-distribution}), this already yields
approximately 343 billion unique UIDs,\footnote{$\sum_{k=5}^{20} \binom{39}{k}
\approx 343$ billion for $N=39$.} far exceeding the user base of any
real-world platform. Domains with more passes provide even greater capacity.
Activating too few passes reduces verification confidence
(\S~\ref{subsec:reliability}), while activating too many increases the collision
probability between users, hindering precise attribution.

\parh{\ding{174} Applying Watermark Passes.}~During inference, when the agent
generates a trajectory for a user with pass set $\mathcal{P}_{id}$, \tool\
intercepts the visible action sequence $\mathbf{a}$. It scans $\mathbf{a}$ to
identify any subsequences that match the equivalence sets $\mathcal{E}_i$ of the
currently active passes in $\mathcal{P}_{id}$. For each identified match, \tool\
samples a replacement segment from the corresponding target distribution
$\hat{D}_i$ and substitutes it in place. The modified trajectory is then
returned to the user.

\subsection{WM Scheme Taxonomy}
\label{subsec:wm-scheme-taxonomy}

\begin{figure}[!ht]
    \centering
    \includegraphics[width=0.95\linewidth]{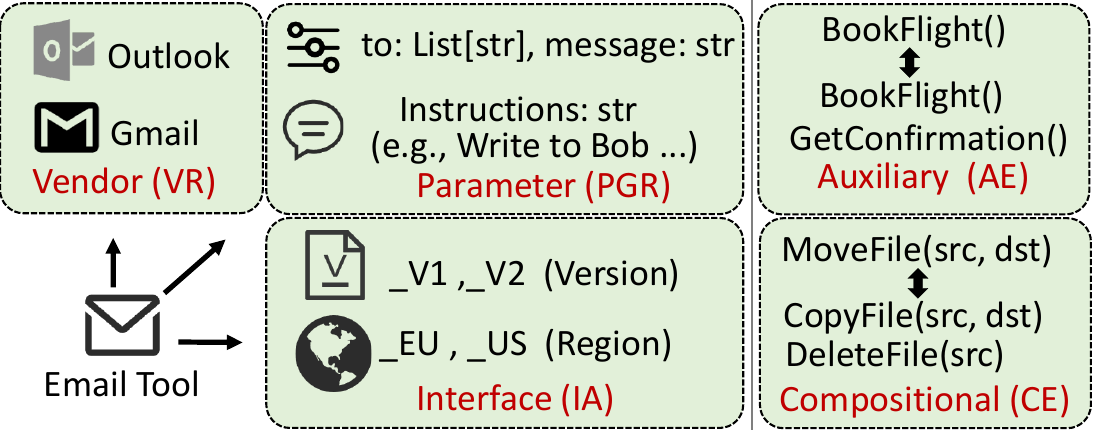}
    \caption{Taxonomy of five watermark schemes. Each defines a distinct
    class of equivalence groups over action segments, enabling
    distribution-level watermark embedding.}
    \label{fig:wm-schemes}
\end{figure}

Each watermark scheme defines a class of equivalence groups over action
segments. Recall that a watermark pass biases the distribution over an
equivalence set $\mathcal{E}_i$ (\S~\ref{subsec:wm-definition}); the schemes
differ in \textit{how equivalence is defined}. Analogous to instruction-level
vs.\ block-level transformations in program analysis, we organize schemes into
two categories (\F~\ref{fig:wm-schemes}). \textit{Action-based} schemes define
equivalence over individual tool calls: each member of $\mathcal{E}_i$ is a
single invocation.
\textit{Structure-based} schemes define equivalence over multi-action
sub-sequences: each member is an ordered sequence of calls that jointly achieve
the same outcome. This orthogonal design expands the combinatorial space an
adversary must search to remove watermarks (\S~\ref{subsec:robustness}). All
schemes preserve task semantics and incur negligible overhead.
\T~\ref{tab:wm-distribution} summarizes the distribution of passes across
schemes and domains.

\parh{Action-Based Schemes.}~Real-world tool ecosystems exhibit pervasive
functional redundancy: multiple tools perform identical operations yet differ
in provider, interface, or parameterization. We identify three orthogonal
sources of such redundancy, each defining a distinct class of equivalence
groups.

\sparh{Vendor Replacement (VR).}~Common functionalities (e.g., email, translation,
weather queries) are served by competing providers. VR groups tools that
provide identical functionality from different vendors into equivalence set.
For instance, $\{w_1 = \texttt{Outlook.SendEmail},\; w_2 = \texttt{Gmail.SendEmail}\}$:
both expose identical semantics, and the vendor choice carries no functional
significance. Such groups arise naturally from ecosystem redundancy, making
the distributional bias inherently stealthy.

\sparh{Parameter Granularity Replacement (PGR).}~Functionally equivalent tools
may expose parameters at different abstraction levels. PGR groups tools that
accept coarser or finer-grained inputs for the same operation. For example, one
email tool takes structured fields (\texttt{to}, \texttt{subject},
\texttt{body}), while another accepts a single natural-language instruction
(e.g., "Write to Bob..."). Both produce the same result. PGR places them in
equivalence set; the interface granularity defines the equivalence.

\sparh{Interface Aliasing (IA).}~Providers often maintain multiple endpoint
names for the same backend (e.g., versioned (\texttt{\_V1} vs.\ \texttt{\_V2}) or
region-specific (\texttt{\_EU} vs.\ \texttt{\_US})). IA groups these
co-referencing names into equivalence set. The alias choice carries zero
semantic information yet produces a detectable distributional signature. Since
aliasing arises from routine API evolution, it blends seamlessly into normal
trajectories.

\parh{Structure-Based Schemes.}~Beyond individual calls, functional
equivalence extends to how actions are \textit{composed}. Structure-based
schemes define equivalence over multi-action sub-sequences that achieve
identical outcomes through different arrangements.

\sparh{Auxiliary Equivalence (AE).}~Agentic models routinely perform ancillary
operations alongside primary actions, such as retrieving confirmation details
or recording metadata. AE defines equivalence between a base sub-sequence and
its variant that includes such ancillary calls. For example,
$[\texttt{BookFlight(route, date)}]$ and
$[\texttt{BookFlight(route, date)}, \texttt{GetConfirmation(id)}]$
form an equivalence group: the flight is booked in both cases, and the
confirmation query merely retrieves an already-assigned reference number
without affecting the reservation. Since agents naturally exhibit both
patterns, the distributional bias remains stealthy.

\sparh{Compositional Equivalence (CE).}~Many operations admit both atomic and
decomposed expressions in tool ecosystems. CE groups these equivalent
representations. For instance, $[\texttt{MoveFile(src, dst)}]$ and
$[\texttt{CopyFile(src, dst)}, \texttt{DeleteFile(src)}]$ form an equivalence
group: both achieve file relocation with identical side effects. The choice
between composite and decomposed forms carries no functional significance,
creating a natural watermark channel.

\subsection{Watermark Instance Generation}
\label{subsec:wm-instance-generation}

While the taxonomy in \S~\ref{subsec:wm-scheme-taxonomy} delineates the
theoretical design space for action equivalence, operationalizing these schemes
requires a scalable generation mechanism. Manual curation of equivalence sets is
prohibitively expensive and inherently unscalable, particularly given the rapid
evolution of real-world tool ecosystems. We therefore construct an automated
pipeline that systematically instantiates watermark passes for the defined
schemes.
Our core design principle is the \textit{decoupling of candidate mining from
equivalence validation}. This separation allows us to maximize recall during
exploration of the vast tool space, while subsequently enforcing precision
through rigorous verification. The pipeline operates in two phases, implemented
by a \textbf{Generator} and a \textbf{Verifier} as shown in
\F~\ref{fig:methodology-overview}(a).

\parh{Phase 1: Candidate Mining (Generator).}~Given a tool library $\mathcal{T}$
and a target watermark scheme $S$, the goal is to identify candidate tool pairs
or sequences that may exhibit equivalence under $S$. We use the
xlam-function-calling-60k dataset~\cite{liu2024apigen} as our tool library
$\mathcal{T}$, comprising 3,673 tools (detailed in \S~\ref{sec:setup}).
To manage the combinatorial explosion of pairwise comparisons, we employ a
two-stage filtering strategy. First, we leverage an embedding Qwen3-embedding
model~\cite{qwenembedding2025} to prune the search space, retaining only tool
pairs whose semantic representations exceed a similarity threshold. This ensures
that only functionally related tools proceed to the next stage. Second, we apply
scheme-specific analysis using an LLM to assess equivalence potential. For
action-based schemes, the LLM examines whether tools perform equivalent
atomic operations (e.g., both compute numerical aggregation). For
structure-based schemes, it analyzes whether tool compositions yield
equivalent execution patterns (e.g., sequential API calls with identical side
effects). This process yields approximately 207 candidate equivalence pairs.
This phase prioritizes recall: we deliberately cast a wide net to capture all
plausible equivalence relationships, accepting false positives for subsequent
refinement in Phase~2.

\parh{Phase 2: Equivalence Validation (Verifier).}~Phase~1 prioritizes recall over
precision, producing candidates that require rigorous verification. We employ a
dual-validation pipeline to eliminate false positives. First, we
invoke a separate LLM to perform semantic analysis, checking whether the
candidate pairs preserve identical functional behavior under representative
inputs. Second, we execute both candidate pairs in our sandboxed
environment~(\S~\ref{sec:setup}) with diverse test cases, comparing outputs and
side effects. Only candidates that pass both semantic and execution validation
are accepted.
Each validated equivalence set $\mathcal{E}_i$ is registered as an operational
watermark pass, expanding the global pool of $101$ passes in
\S~\ref{subsec:wm-insertion-procedure}. This automated pipeline produces
hundreds of verified watermark instances with minimal manual effort. The first
two authors further conduct post-hoc human inspection on a random sample of 50
validated instances (half of the total) and observe no functional
inconsistencies, confirming the reliability of our automated validation
pipeline.

\section{IP Theft Verification Stage}
\label{sec:check}

In this section, we detail the verification framework used to identify IP theft.
The procedure moves from individual WM detection to holistic model
classification and, finally, attacker localization.

\subsection{Motivation and Assumptions}
\label{subsec:check-assumptions}

\parh{Motivation.}~The owner of the victim model $\mathcal{M}_{vic}$ seeks to
verify whether a suspect model $\mathcal{M}_{imi}$ was trained on its
watermarked outputs. As illustrated in \F~\ref{fig:methodology-overview}, this verification exploits the
principle that imitation learning preserves the statistical biases injected by
\tool into the action trajectories. If $\mathcal{M}_{imi}$ was fine-tuned on
watermarked trajectories, it will exhibit the same distributional patterns
embedded during watermark injection.

\parh{Assumptions.}~Following the threat model defined in
\S~\ref{subsec:threat-model}, the adversary leverages grey-box trajectories to
fine-tune a public backbone $\mathcal{M}_{pub}$. Each registered user receives a
unique identifier (UID) $U_{id}$ mapped to a distinct WM pass set
$\mathcal{P}_{id}$, enabling attribution of theft to specific malicious users.
Regarding watermark knowledge, we consider two levels of adversary awareness. A
\textbf{Partially Knowledgeable (PK) Attacker} is aware of the watermark's
existence but lacks knowledge of the specific watermarking schemes employed by
\tool. A \textbf{Fully Knowledgeable (FK) Attacker} knows all five candidate
watermarking schemes but cannot determine the exact WM pass set
$\mathcal{P}_{id}$ assigned to the target service. Each watermarking scheme
involves numerous configurable parameters (such as which actions to transform,
the specific transformation rules, and watermark strength), and the space of
possible configurations grows exponentially. Consequently, even with full
knowledge of the schemes, the FK attacker faces substantial uncertainty in
identifying or removing the precise watermark configuration.

For verification, the owner uses a private dataset $\mathcal{V}_{v}$ to query
$\mathcal{M}_{imi}$. Following standard assumptions in prior watermarking
literature~\cite{li2023protecting}, we assume $\mathcal{V}_{v}$ is not publicly accessible and
shares the same distribution as the original fine-tuning data $D_{\text{ft}}$
used to train $\mathcal{M}_{vic}$.

\subsection{Verification Procedure}
\label{subsec:verification-procedure}

\parh{Detection of Individual WM Passes.}~To check for a specific WM pass $P_i$,
we query the suspect model $\mathcal{M}_{imi}$ with the private verification
dataset $\mathcal{V}_{v}$ and collect its grey-box trajectories. From these
trajectories, we compute the empirical distribution $D'_i$ of equivalent action
segment sets. We then quantify the discrepancy between $D'_i$ and the target
distribution $\hat{D}_i$ embedded during watermark injection using
Jensen-Shannon Divergence (JSD). As a symmetric metric bounded in $[0, 1]$, a
lower JSD value signifies greater distributional similarity. A pass $P_i$ is
considered detected if $\text{JSD}(D'_i, \hat{D}_i) < \theta_{J}$, where
$\theta_{J}$ is the sensitivity threshold. A JSD value below $\theta_{J}$
provides strong evidence that the training data of $\mathcal{M}_{imi}$ contains
the watermark pattern from pass $P_i$.

\parh{Holistic Model Classification.}~Since the identity of the model's trainer
is initially unknown during verification, we test $\mathcal{M}_{imi}$ against
the global pool of all $N$ WM passes in \tool. For each pass $P_i$, we mark it
as detected if its JSD value falls below $\theta_{J}$. Let $n_{det}$ denote the
total count of detected passes. The suspect model $\mathcal{M}_{imi}$ is
classified as an imitation model only if $n_{det} \geq \theta_{N}$, where
$\theta_{N}$ is the detection threshold. This two-level thresholding design
balances robustness and accuracy: $\theta_{J}$ controls per-pass sensitivity to
reduce noise, while $\theta_{N}$ requires multiple passes to confirm theft,
thereby minimizing false positives. We evaluate the impact of these thresholds
on verification accuracy in \S~\ref{subsec:reliability}. Most WM passes operate independently,
ensuring that one watermark does not interfere with another. However, a few
passes may affect the transformable candidates of subsequent passes. To ensure
consistency, we apply these interdependent passes in a fixed order during both
injection and verification, as defined in \S~\ref{subsec:wm-insertion-procedure}.

\parh{Localization of the Malicious User.}~Once theft is confirmed, \tool
identifies the specific attacker by exploiting the unique mapping between
UIDs and pass sets established during user registration (\S~\ref{subsec:wm-insertion-procedure}). We compare
the binary vector of detected passes in $\mathcal{M}_{imi}$ against the registry
of all assigned WM pass sets. To account for potential signal loss during
imitation learning and data sampling variance, we employ similarity-based
matching (e.g., cosine similarity) rather than requiring exact
matches. A malicious user is identified when their assigned WM pass
set exhibits the highest similarity to the detected passes in
$\mathcal{M}_{imi}$. We evaluate the accuracy and robustness of this
localization procedure in \S~\ref{subsec:reliability}, including scenarios where
one or two WM passes fail detection due to imitation learning quality or
statistical noise.
\section{Implementation and Setup}
\label{sec:setup}

\begin{table}[!t]
\centering
\caption{Distribution of watermark passes across five schemes and three domains.}
\label{tab:wm-distribution}
\resizebox{0.98\linewidth}{!}{
\begin{tabular}{lccccc}
\toprule
\multirow{2}{*}{Scheme} & \multicolumn{3}{c}{Domain} & \multirow{2}{*}{Total} \\
\cmidrule(lr){2-4}
 & Data & Business & Social &  \\
\midrule
Vendor Replacement (VR) & 8 & 12 & 18 & 38 \\
Parameter Granularity Replacement (PGR) & 7 & 4 & 2 & 13 \\
Interface Aliasing (IA) & 11 & 5 & 6 & 22 \\
\midrule
Auxiliary Equivalence (AE) & 7 & 5 & 3 & 15 \\
Compositional Equivalence (CE) & 6 & 2 & 5 & 13 \\
\midrule
Total & 39 & 28 & 34 & 101 \\
\bottomrule
\end{tabular}
}
\end{table}

We now present the experimental setup. All experiments are performed with four
NVIDIA H800 graphics cards.

\parh{Dataset.}~We build upon xlam-function-calling-60k~\cite{liu2024apigen},
which contains 60,000 trajectories covering 3,673 tools across 21 domains. Since
the original dataset contains limited multi-step trajectories, following the
original methodology, we extend it and validate the correctness of each
generated trajectory using GPT-5~\cite{openai2023} and
DeepSeek-v3.2~\cite{liu2025deepseek}. 
We select three high-impact domains (i.e., Social, Business, and Data) based on
their prevalence in real-world agentic platforms~\cite{coze2025,dify2025} and
prior work~\cite{yao2024tau}. The curated dataset comprises 30,000 trajectories
in total. We split them into a fine-tuning set $D_{\text{ft}}$ (24,000 samples) and a
held-out test set $D_{\text{test}}$ (6,000 samples), following standard watermarking
protocols~\cite{li2023protecting}.
We use $D_{\text{ft}}$ to simulate the attack, representing an upper-bound
threat where adversaries possess a diverse query set. Although real attackers
often bootstrap from seed queries~\cite{li2025differentiation}, this setting
provides a conservative evaluation of watermark robustness.
We evaluate $\mathcal{M}_{imi}$ on $D_{\text{test}}$ and use $D_{\text{test}}$
as the verification set for IP theft detection, consistent with prior
work~\cite{li2023protecting}.

\parh{Metrics.}~Following standard evaluation protocols for agentic
systems~\cite{qintoolllm,chang2024agentboard,zhugeagent}, we assess model
performance at both task and trajectory levels. For final response quality,
we measure \emph{Pass Rate} (PR) and \emph{Response Score} (RS). PR is
a binary metric determining if the response fulfills the user request, while RS
evaluates semantic quality on a 1--5 scale. For trajectory quality, we adopt
\emph{Trajectory Score} (TS)~\cite{chang2024agentboard,zhugeagent}, which
evaluates the correctness and coherence of the reasoning chain on a 1--5 scale.
All three metrics are scored by DeepSeek-v3.2 acting as a judge.
To enable fair comparison, we normalize all scores to the [0, 1] range.
We also measure
downstream capabilities on ARC~\cite{clark2018think} for reasoning and
ACPBench~\cite{kokel2025acpbench} for planning. For watermark reliability, we
measure Precision, Recall, and F1-score for detection, and Top-1 Accuracy for
malicious user localization~\cite{li2023protecting}. To assess stealth, we measure
perplexity (PPL)~\cite{alon2023detecting}, which quantifies how much the watermark
shifts output distribution from its natural behavior.

\parh{Model \& Agentic System.}~We employ Qwen3-4B~\cite{yang2025qwen3} as the
base public backbone $\mathcal{M}_{pub}$ in our main experiments, a widely
adopted choice in recent agentic research~\cite{kang2025distilling}. We further
evaluate generalization on Qwen3-8B~\cite{yang2025qwen3} and
Ministral-8B~\cite{mistral2024} in \S~\ref{subsec:ablation}. For the agentic
framework, we adopt ReAct~\cite{yao2022react}. Following~\cite{qintoolllm}, we
use Qwen3-Next-80B-A3B~\cite{qwen3next2025} as the environment simulator to
execute tools and generate observations. We fine-tune all models using
LoRA~\cite{hu2022lora} with rank $r=64$, alpha $\alpha=128$, and a learning rate
of $2e-4$.

\parh{Watermark Configuration.}~We instantiate a total of 101 watermark
passes across 5 schemes (\S~\ref{subsec:wm-scheme-taxonomy}), with $N$ ranging
from 28 to 39 per domain (see \T~\ref{tab:wm-distribution}). Within each domain,
each user is assigned a unique $N$-bit User ID ($U_{id}$), where the $i$-th bit
corresponds to the activation of the $i$-th watermark pass. For each user, we
randomly activate a subset of passes (5--20) to ensure high user capacity while
preserving utility.

\section{Evaluation}
\label{sec:evaluation}

To assess whether \tool\ meets the design properties outlined in
\S~\ref{subsec:design-considerations}, we organize the evaluation around five research questions (RQs):

\parh{RQ1: Fidelity \& Cost.}~Does watermark injection degrade the functional
quality of protected agentic systems? How much computational overhead does
\tool\ introduce during inference?

\parh{RQ2: Reliability.}~How accurately can \tool\ identify the source of
IP theft when performing verification on a suspect agentic system?

\parh{RQ3: Robustness.}~Can \tool's watermarks withstand removal attacks that
attempt to erase embedded signatures?

\parh{RQ4: Stealthiness.}~Are watermarked outputs distinguishable from
clean outputs to human observers or automated detectors?

\parh{RQ5: Ablation Study.}~How does \tool\ generalize across different backbone
LLMs, watermark strengths, and cross-model attack scenarios?

\subsection{RQ1: Fidelity \& Cost}
\label{subsec:fidelity}

We evaluate whether watermark injection degrades model quality for both
legitimate users and adversaries. Given the grey-box visibility constraint
(\S~\ref{subsec:threat-model}), we assess trajectory fidelity indirectly by
measuring $\mathcal{M}_{imi}$'s performance: if watermarks corrupt trajectories,
models trained on them would exhibit severe degradation~\cite{li2023protecting}.
We evaluate three aspects: (1) trajectory quality (PR, RS, TS) on
domain-specific tasks, (2) downstream capabilities (ARC, ACPBench) to verify
general abilities remain intact, and (3) computational overhead during
inference. For $\mathcal{M}_{imi}$, we test three watermark intensities:
5--10, 10--15, and 15--20 passes.

\begin{table}[!t]
    \centering
    \caption{Fidelity metrics (PR, RS, TS) of victim and imitated models across different domains.}
    \label{tab:fidelity}
    \resizebox{\linewidth}{!}{
    \begin{tabular}{ccccccccccc}
    \toprule
    \multicolumn{2}{c}{Domain} & \multicolumn{3}{c}{Data} &
    \multicolumn{3}{c}{Business} & \multicolumn{3}{c}{Social} \\
    \cmidrule(lr){3-5} \cmidrule(lr){6-8} \cmidrule(lr){9-11}
    \multicolumn{2}{c}{Metric} & PR & RS & TS & PR & RS & TS & PR & RS & TS
    \\
\midrule
\multicolumn{2}{c}{$\mathcal{M}_{vic}$} &0.792 & 0.630 & 0.896 & 0.687& 0.574& 0.903&0.707 & 0.582 & 0.943\\
\midrule
\multirow{3}{*}{$\mathcal{M}_{imi}$} & 5--10 WM passes &0.755 & 0.575 & 0.837 & 0.642& 0.543& 0.841&0.678 & 0.529 & 0.871\\
&10--15 WM passes &0.703 & 0.591 & 0.860 & 0.637& 0.537& 0.858& 0.663& 0.536 & 0.868\\
& 15--20 WM passes &0.741 & 0.586 & 0.841 & 0.664& 0.553& 0.834&0.626 & 0.539 & 0.884\\
    \bottomrule
    \end{tabular}
    }
    \end{table}

\parh{Final Response and Trajectory Quality.}~We find that watermark
injection preserves both victim and imitation model quality. As shown in
\T~\ref{tab:fidelity}, $\mathcal{M}_{imi}$ trained on watermarked trajectories
maintains comparable performance to $\mathcal{M}_{vic}$ across all domains.
For instance, in the Data domain, $\mathcal{M}_{imi}$ achieves PR of 0.703--0.755
compared to $\mathcal{M}_{vic}$'s 0.792. This demonstrates two critical properties.
First, watermarking does not corrupt the functional quality of trajectories, as
$\mathcal{M}_{imi}$ successfully learns domain capabilities from them. Second,
the consistent performance across varying watermark intensities (5--10,
10--15, 15--20 passes) shows that \tool\ achieves a practical
balance: strong enough watermarks for reliable verification (\S~\ref{subsec:reliability}) without degrading
utility. The slight performance gap between $\mathcal{M}_{vic}$ and
$\mathcal{M}_{imi}$ arises from inherent limitations in grey-box trajectory
reconstruction, not from watermark corruption, confirming that watermarked
trajectories remain valuable training signals for adversaries.

\parh{Downstream Performance.}~To rule out hidden degradation outside protected domains, we evaluate both
$\mathcal{M}_{vic}$ and $\mathcal{M}_{imi}$ on ARC (complex reasoning) and
ACPBench (agent planning).
As shown in \T~\ref{tab:fidelity2}, watermarked models preserve downstream 
performance across all domains. For instance, in the Data domain, 
$\mathcal{M}_{imi}$ achieves 80.09\% on ARC compared to $\mathcal{M}_{vic}$'s 
80.43\%, demonstrating negligible degradation. This holds across different 
watermark intensities and benchmarks, confirming that \tool's trajectory-level 
watermarking operates orthogonally to the model's general reasoning capabilities.
This result addresses a critical concern: watermarking only affects trajectory 
selection within the protected domain, leaving the model's foundational abilities 
intact. Combined with our trajectory quality results (\T~\ref{tab:fidelity}), 
this confirms that \tool\ achieves complete fidelity preservation across both 
domain-specific and general-purpose capabilities.

\begin{table}[!t]
    \centering
    \caption{Downstream task performance of victim and imitated models across different domains.}
    \label{tab:fidelity2}
    \resizebox{\linewidth}{!}{
    \begin{tabular}{cccccccc}
    \toprule
    \multicolumn{2}{c}{Domain} & \multicolumn{2}{c}{Data} &
    \multicolumn{2}{c}{Business} & \multicolumn{2}{c}{Social} \\
    \cmidrule(lr){3-4} \cmidrule(lr){5-6} \cmidrule(lr){7-8}
    \multicolumn{2}{c}{Downstream task} & ARC & ACPBench & ARC & ACPBench & ARC & ACPBench \\
    \midrule
    \multicolumn{2}{c}{$\mathcal{M}_{vic}$} & 80.43\% & 46.69\% & 79.92\% & 59.15\% & 80.60\% & 59.62\% \\
    \midrule
    \multirow{3}{*}{$\mathcal{M}_{imi}$} & 5--10 passes & 79.88\% & 55.30\% & 79.71\% & 55.68\% & 79.12\% & 57.18\% \\
    & 10--15 passes & 80.09\% & 49.72\% & 79.63\% & 55.21\% & 79.29\% & 55.83\% \\
    & 15--20 passes & 81.06\% & 50.66\% & 80.01\% & 58.91\% & 78.91\% & 56.00\% \\
    \bottomrule
    \end{tabular}
    }
    \end{table}
    
    \begin{table}[!t]
        \centering
        \caption{Overhead of watermark insertion over 1,000 queries.}
        \label{tab:cost}
        \resizebox{0.95\linewidth}{!}{
        \begin{tabular}{ccc}
        \toprule
        Response Time (s) & WM Insertion Time (s) & Response Slowdown  \\
        \midrule 
        3428 & 9.6 &  0.28\% \\
        \bottomrule
        \end{tabular}
        }
        \end{table}

\parh{Cost.}~Beyond fidelity, we quantify the computational overhead 
introduced by watermark injection. We measure the additional latency across 
1,000 queries sampled from $D_{\text{test}}$ on $\mathcal{M}_{vic}$.
As shown in \T~\ref{tab:cost}, watermark injection introduces only 9.6 seconds 
of overhead across all 1,000 queries, corresponding to a negligible 0.28\% 
slowdown in total response time. This minimal overhead stems from \tool's 
design: watermark injection occurs during trajectory generation 
(\S~\ref{sec:methodology}), where we only manipulate action sequences
 without invoking additional LLM calls or external computations. The 
overhead primarily consists of lightweight pass activation checks.
This result demonstrates that \tool\ achieves practical deployability: 
watermark protection incurs virtually no performance penalty, making it viable 
for production agentic systems serving real-time user queries.

\subsection{RQ2: Reliability}
\label{subsec:reliability}

To evaluate reliability, we assess whether \tool\ can accurately (1) identify suspect models $\mathcal{M}_{imi}$ trained on watermarked trajectories and (2) pinpoint the specific malicious user (UID) responsible for the attack.

\parh{Confirming a Suspect Model.}~As described in \S~\ref{sec:check}, given a
suspect model $\mathcal{M}_{imi}$, we query it with a private verification
dataset $\mathcal{V}_{v}$ to determine whether $\mathcal{M}_{imi}$ was trained
on watermarked trajectories stolen from $\mathcal{M}_{vic}$. Following prior
watermarking work~\cite{li2023protecting}, we use the test split of $D_{\text{test}}$ as
$\mathcal{V}_{v}$ in our evaluation.

To ensure statistical reliability within a feasible computational budget
(1,000 GPU hours), we fine-tuned a total of 36 imitation models
across the three domains.
For each domain, we leverage 12 of these models as positive samples (i.e.,
models trained on stolen watermarked trajectories). To evaluate false positive
rates, we also prepare 12 benign models as negative samples per domain. 
These
benign models are randomly sampled from the clean $D_{\text{ft}}$ and
fine-tuned without any watermark injection. This setup ensures realistic
evaluation of both detection accuracy and false alarm rates.

\begin{figure}[!t]{
    \centering
    \includegraphics[width=\linewidth]{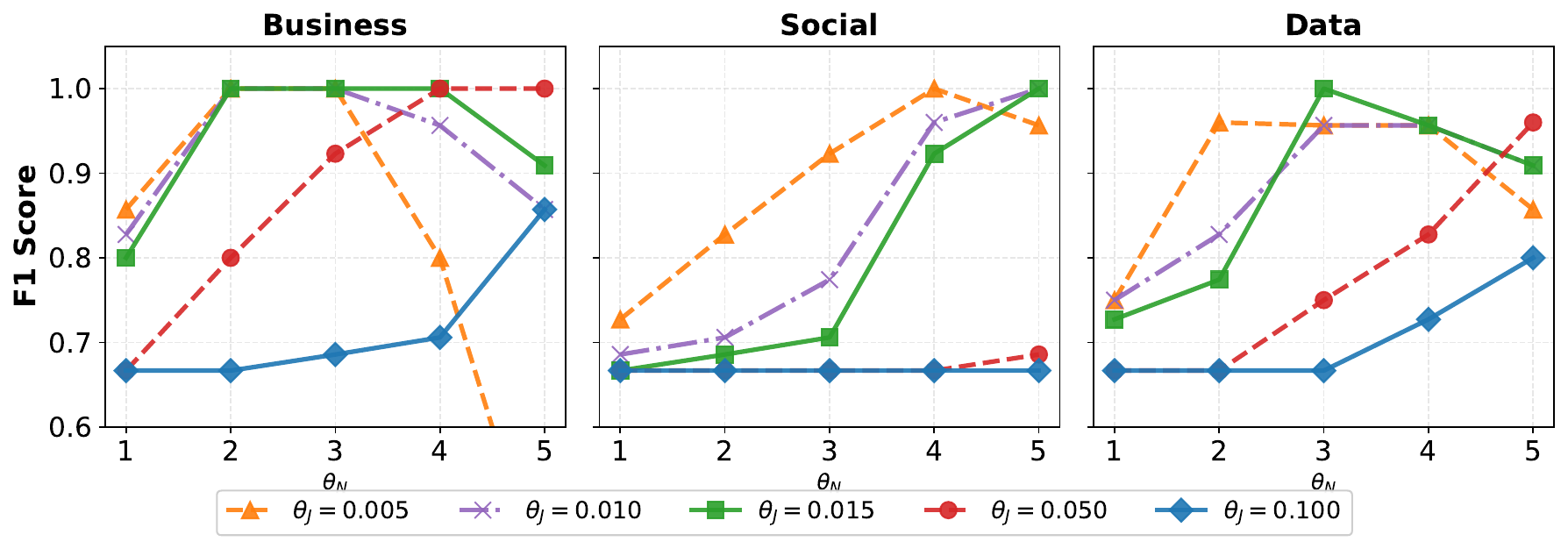}
    \caption{Detection performance across different threshold configurations.}
    \label{fig:f1}
    }
    \end{figure}

Our verification process (\S~\ref{sec:check}) relies on two hyper-parameters:
$\theta_{J}$ (the JSD threshold for detecting individual watermark passes) and
$\theta_{N}$ (the minimum number of detected passes required to confirm IP
theft). To systematically evaluate detection accuracy, we test five JSD
thresholds: $\theta_{J} \in \{0.005, 0.010, 0.015, 0.050, 0.100\}$. For
$\theta_{N}$, we consider values from 1 to 5, reflecting the fact that we inject
at least 5 watermark passes per user (\S~\ref{subsec:fidelity}).
We evaluate all configurations across three domains and measure performance
using the F1-score: $F_1 = \frac{2PR}{P+R}$, where precision $P$ captures the
fraction of correctly identified stolen models among all flagged models, and
recall $R$ measures the fraction of stolen models successfully detected. Results
are shown in \F~\ref{fig:f1}.

\F~\ref{fig:f1} demonstrates that \tool\ achieves reliable IP theft detection across diverse scenarios. With appropriate threshold configurations, \tool\ consistently delivers high F1-scores (nearly 1.00) across
all three domains, confirming its effectiveness in identifying watermarked
trajectories. However, detection accuracy degrades when $\theta_{N}$ is improperly configured. When $\theta_{N}$ is too low (e.g., $\theta_{N}=1$ or $2$), the system becomes overly sensitive, flagging benign models that coincidentally
trigger a single watermark pass. Conversely, excessively high $\theta_{N}$
values (e.g., $\theta_{N} \geq 4$) reduce detection power, as attackers may not
collect enough queries to activate all embedded watermark passes. Based on our
empirical analysis, we recommend setting $\theta_{N}=3$ as a practical default,
which balances robustness against false positives while maintaining sufficient
detection sensitivity.

\parh{Localizing the Malicious User.}~Beyond confirming IP theft, we evaluate a
harder task: pinpointing the specific malicious user from a large pool.
Given 12 known malicious imitation models, we first identify each model's
activated watermark passes via statistical testing. The activated passes form a
binary fingerprint, represented as an $N$-bit vector $\mathbf{v}$, where $N$ is
the number of active passes in the domain. We then match $\mathbf{v}$ against
the ground-truth watermark assignments $\mathbf{p}_i$ for all users using cosine
similarity, returning the highest-similarity user as the identified adversary.

\T~\ref{tab:user-localization} reports results across varying pool sizes: 12
malicious users alone, and 12 malicious users mixed with 1K, 2K, 5K, and 50K
randomly sampled benign users. Each experiment is repeated multiple times and we
report the averaged top-1 accuracy. \tool\ maintains high localization accuracy
across all domains. Data and Business achieve 0.92--1.0 accuracy even at 50K
total users. The Social domain exhibits lower accuracy (0.81 at 50K) because its
tool ecosystem yields fewer distinguishing passes per trajectory, reducing
fingerprint uniqueness. Overall, the accuracy degradation at extreme scales
stems from increased watermark collision probability among benign users, not
fundamental design limitations.

While \tool\ theoretically supports nearly 343 billion unique identities through
combinatorial watermark assignments (5--20 passes), exhaustive search is
unnecessary. Model extraction attacks require extensive querying, so
investigators can filter by query volume (e.g., $>10$K queries), narrowing the
search space from millions to a few thousand suspects.

\begin{table}[!t]
    \centering
    \caption{Adversary localization accuracy under different user pool sizes with top-1 matching.}
    \label{tab:user-localization}
    \resizebox{0.95\linewidth}{!}{
    \begin{tabular}{lccccc}
    \toprule
    \multirow{2}{*}{Domain} & \multicolumn{5}{c}{Total User Pool Size} \\
    \cmidrule(lr){2-6}
     & 12 & 12 + 1K & 12 + 2K & 12 + 5K & 12 + 50K \\
    \midrule
    Data & 1.0 & 1.0 & 0.97 & 0.97 & 0.92 \\
    Business & 1.0 & 1.0 & 1.0 & 0.97 & 0.94 \\
    Social & 0.92 & 0.92 & 0.89 & 0.86 & 0.81 \\
    \bottomrule
    \end{tabular}
    }
    \end{table}

\subsection{RQ3: Robustness}
\label{subsec:robustness}

Robustness measures whether \tool's watermarks survive adversarial removal
attempts. The core of our defense lies in asymmetric uncertainty: while
adversaries may know watermarking exists, the vast design space and flexible
instantiation of our five-scheme framework make targeted removal extremely
difficult. Following the threat levels defined in
\S~\ref{subsec:check-assumptions}, we evaluate both \textit{Partially
Knowledgeable} (PK) and \textit{Fully Knowledgeable} (FK) attackers.

\parh{PK Attackers.}~PK attackers operate in a black-box setting: they know 
watermarks exist but lack knowledge of the five schemes or their instantiation. 
We assume they deploy generic removal strategies adapted from text watermarking 
literature~\cite{chenmark,zhang2024remark}. We evaluate three representative attacks.
\textit{Random Deletion} randomly removes tokens
from trajectories to disrupt embedded patterns. \textit{LLM
Rephrase} instructs an LLM to paraphrase
trajectories while preserving the quality, aiming to erase watermark signals
through linguistic variation. \textit{LLM
Replacement} leverages an LLM to identify and
substitute potentially watermarked tokens with natural alternatives, balancing
removal effectiveness and trajectory quality.
We evaluate watermark attack using 1K watermarked trajectories generated by
\tool. For each attack strategy, we measure watermark attack accuracy via precision
(fraction of detected tokens that are true watermarks), recall (fraction of true
watermarks successfully detected), and F1-score (harmonic mean of precision and
recall). \T~\ref{tab:wm-recognition} shows results.

\T~\ref{tab:wm-recognition} shows that all three attacks fail to remove
watermarks effectively. All F1-scores remain below 0.02, indicating attackers
can barely identify watermark-carrying tokens. This failure stems from a
fundamental mismatch: these attacks target token-level patterns inherited from
text watermarking, whereas \tool\ embeds watermarks into action-sequence
distributions at the trajectory level. Without knowledge of our five schemes or
their instantiation, attackers cannot distinguish watermarked actions from
natural variations in agentic reasoning. The low recall values (e.g., 0.0517 for
LLM Rephrase) confirm that generic removal strategies remain blind to our
distribution-level watermarks.

\begin{table}[!t]
\centering
\caption{Watermark detection under different attacks.}
\label{tab:wm-recognition}
\resizebox{0.8\linewidth}{!}{
\begin{tabular}{lccc}
\toprule
\textbf{Attack Strategy} & \textbf{Precision} & \textbf{Recall} & \textbf{F1-Score} \\
\midrule
Random Deletion & 0.0056 & 0.0917 & 0.0103 \\
LLM Rephrase & 0.0012 & 0.0517 & 0.0023 \\
LLM Replacement & 0.0020 & 0.1817 & 0.0040 \\
\midrule
FK Replacement & 0.2644 & 0.2256 & 0.2350\\
\bottomrule
\end{tabular}
}
\end{table}

\parh{FK Attackers.}~As mentioned before, FK attackers represent the worst-case threat: they 
know all five watermark schemes but face a combinatorial challenge. Each 
scheme can generate many distinct passes (\S~\ref{subsec:wm-scheme-taxonomy}), 
and without knowing which specific passes were activated for a given user, 
attackers cannot identify watermarked action sequences in collected 
trajectories. This uncertainty forces them to guess which actions carry 
watermark signals. To evaluate this worst-case scenario, we design an FK Replacement attack that 
exploits their scheme knowledge to optimize the baseline LLM Replacement 
strategy. This simulates their best effort at identifying and removing 
watermarked actions.

\begin{table}[!h]
    \centering
    \caption{Trajectory quality degradation after FK attackers 
    attempt watermark removal. Since the final response remains unchanged, 
    the pass rate is identical; thus we only report RS and TS.}
    \label{tab:fs-results}
    \resizebox{0.6\linewidth}{!}{
    \renewcommand{\arraystretch}{0.7}
    \begin{tabular}{ccc}
    \toprule
    Metric  & RS & TS \\
    \midrule
    Original   & 0.612 & 0.877 \\
    Watermark Removed & 0.391 & 0.514 \\
    \bottomrule
    \end{tabular}
    }
\end{table}

Results appear in the last row of \T~\ref{tab:wm-recognition}.
With scheme knowledge, both recall and
precision improve substantially over the PK scenario. However,
the F1-score (0.2350) remains low. 
Moreover, removal attempts severely degrade trajectory quality.
As \T~\ref{tab:fs-results} shows, RS drops from 0.612 to 0.391 after removal,
because FK attackers cannot pinpoint which passes were activated and
must over-modify action sequences, disrupting the reasoning coherence.
Thus, even with full scheme knowledge, attackers cannot remove \tool's watermark
without rendering the stolen data unusable.

\subsection{RQ4: Stealthiness}
\label{subsec:stealthiness}

Stealthiness measures whether attackers can distinguish watermarked 
trajectories from clean ones. If adversaries detect watermark patterns during 
data collection, they can filter out watermarked samples before fine-tuning 
$\mathcal{M}_{imi}$, rendering the watermark ineffective. A stealthy watermark 
must preserve the statistical properties of natural trajectories, making 
detection infeasible without access to the victim's private verification 
procedure.

Following prior work~\cite{wangtowards}, we measure stealthiness using
perplexity (PPL). PPL is computed by an external language model (GPT-2~\cite{radford2019language})
 to quantify text fluency,
where lower values indicate more natural text. We compare the PPL of
trajectories before and after watermark injection to quantify the deviation
introduced by embedded watermarks.

\begin{figure}[!htbp]{
\centering
\includegraphics[width=0.95\linewidth]{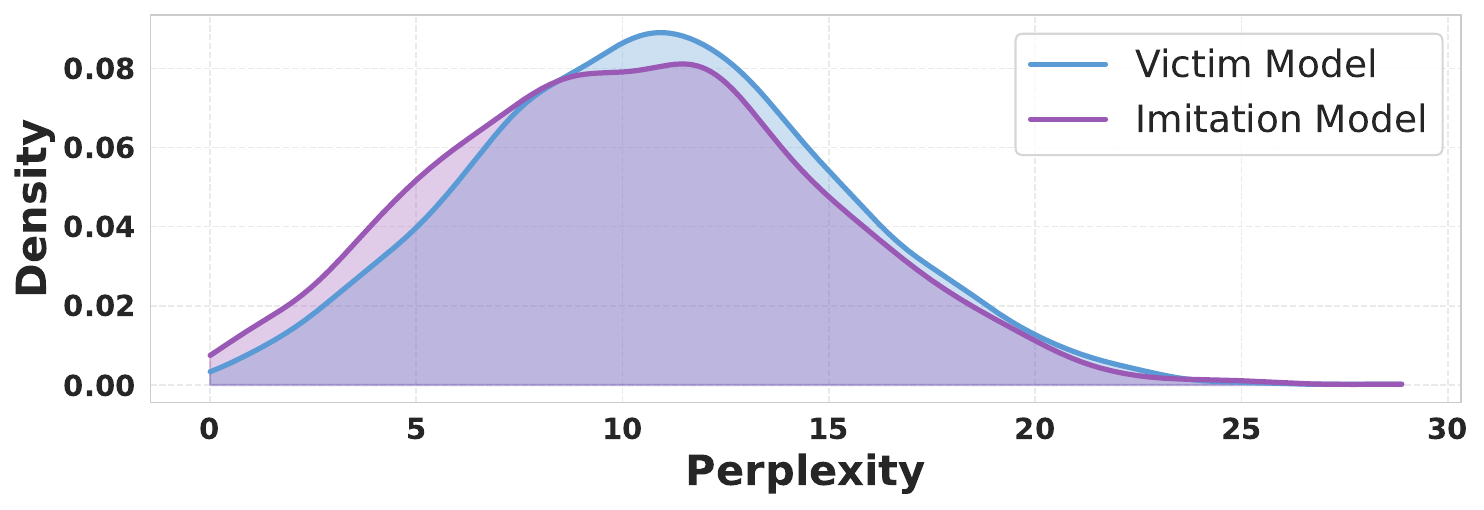}
\caption{PPL comparison between $\mathcal{M}_{vic}$ and $\mathcal{M}_{imi}$.}
\label{fig:perplexity}
}
\end{figure}

\F~\ref{fig:perplexity} visualizes the PPL densities. The distributions exhibit a near-perfect overlap, with the difference in mean PPL being less than 0.5. This statistical indistinguishability confirms that \tool's distribution biasing is subtle enough to remain hidden within the inherent variance of agentic reasoning.
The result validates our design principle: by manipulating action-sequence 
distributions rather than individual tokens, \tool\ preserves the natural 
fluency of agentic reasoning. From an adversary's perspective, watermarked 
trajectories remain indistinguishable from clean ones using automated 
perplexity-based detection. 
This stealthiness property is critical: it prevents attackers from filtering 
watermarked samples during data collection, ensuring that fine-tuning on 
stolen trajectories inevitably inherits embedded watermark signals. Combined 
with our fidelity results (\S~\ref{subsec:fidelity}), this demonstrates that 
\tool\ achieves invisible protection without sacrificing utility.

\subsection{RQ5: Ablation Study}
\label{subsec:ablation}

\parh{Impact of Backbone LLM.}~We evaluate whether \tool\ generalizes
beyond Qwen3-4B by extending to Ministral-8B~\cite{mistral2024} and
Qwen3-8B~\cite{yang2025qwen3}, which differ in model family and scale.
We retrain both $\mathcal{M}_{vic}$ and $\mathcal{M}_{imi}$ from scratch
for each backbone. \T~\ref{tab:llm-comparison} reports the results.
\tool\ maintains consistent performance across all backbones.
Both Ministral-8B and Qwen3-8B achieve fidelity comparable to
Qwen3-4B (\S~\ref{subsec:fidelity}), with only marginal degradation
from $\mathcal{M}_{vic}$ to $\mathcal{M}_{imi}$ (e.g., PR: 0.742
$\to$ 0.678 on Qwen3-8B). Reliability also holds: top-1 accuracy
remains 0.93--1.0 under 5K--50K user pools.
This backbone-agnostic behavior follows from \tool's design:
watermarks are defined over action-sequence distributions
(\S~\ref{subsec:wm-definition}), not model-specific embeddings,
making them invariant to the underlying architecture. In practice,
providers can swap backbone LLMs without re-engineering the
watermarking pipeline.

\begin{table}[!t]
    \centering
    \caption{Performance comparison across different LLM.}
    \label{tab:llm-comparison}
    \resizebox{0.84\linewidth}{!}{
    \begin{tabular}{cccccccc}
    \toprule
    \multicolumn{2}{c}{Model} & \multicolumn{3}{c}{Fidelity} & \multicolumn{2}{c}{Reliability} \\
    \cmidrule(lr){3-5} \cmidrule(lr){6-7}
    Type &  & PR & RS & TS & 5K & 50K  \\
    \midrule
    \multirow{2}{*}{Ministral} & $\mathcal{M}_{vic}$ & 0.702 & 0.582 & 0.904& -- & -- \\
    & $\mathcal{M}_{imi}$ & 0.633 & 0.539 & 0.846 & 0.93 & 0.87 \\
    \midrule
    \multirow{2}{*}{Qwen 8B} & $\mathcal{M}_{vic}$ & 0.742 & 0.564 & 0.910 & -- & -- \\
    & $\mathcal{M}_{imi}$ & 0.678 & 0.581 & 0.878 & 1.0 & 0.93 \\
    \bottomrule
    \end{tabular}
    }
\end{table}

\parh{Impact of Cross-model Imitation.}~Prior experiments assume both
victim and attacker share the same backbone LLM. In practice, adversaries
may fine-tune a different public model (\S~\ref{subsec:threat-model}).
We evaluate this scenario by building $\mathcal{M}_{vic}$ on Ministral-8B
or Qwen3-8B, while $\mathcal{M}_{imi}$ uses Qwen3-4B fine-tuned on
stolen trajectories.
As shown in \T~\ref{tab:backbone-comparison}, cross-model imitation
remains effective. When the victim uses Ministral-8B, the Qwen3-4B-based
$\mathcal{M}_{imi}$ achieves a PR of 0.661 versus the victim's 0.702.
Despite differing architectures and fewer parameters,
$\mathcal{M}_{imi}$ closely replicates the victim's capabilities,
confirming that imitation threats extend beyond same-backbone scenarios.

\begin{table}[!t]
    \centering
    \caption{Cross-model imitation performance.}
    \label{tab:backbone-comparison}
    \resizebox{0.75\linewidth}{!}{
    \begin{tabular}{lccc}
    \toprule
    Model & PR & RS & TS \\
    \midrule
    $\mathcal{M}_{vic}$ (Ministral-8B) & 0.702 & 0.582 & 0.904 \\
    $\mathcal{M}_{imi}$ (Qwen3-4B) & 0.661 & 0.562 & 0.858 \\
    \midrule
    $\mathcal{M}_{vic}$ (Qwen3-8B) & 0.742 & 0.586 & 0.910 \\
    $\mathcal{M}_{imi}$ (Qwen3-4B) & 0.689 & 0.565 & 0.873 \\
    \bottomrule
    \end{tabular}
    }
\end{table}

\begin{table}[!ht]
    \centering
    \caption{Impact of watermark strength ($\delta$).}
    \label{tab:delta-impact}
    \resizebox{0.95\linewidth}{!}{
    \begin{tabular}{ccccccc}
    \toprule
    Watermark Strength ($\delta$) & 0 & 1 & 2 & 3 & 4 & 5 \\
    \midrule
    KLD&-- &0.096 & 0.386& 0.801& 1.408& 1.808\\
    TS & 0.727&0.712& 0.712 &0.707 &0.694 & 0.671 \\
    \bottomrule
    \end{tabular}
    }
\end{table}

\parh{Impact of $\delta$ on Watermark Strength.}~As mentioned in
\S~\ref{subsec:wm-insertion-procedure}, $\delta$ controls the bias strength in
our target distribution derivation. To understand how $\delta$ affects both
watermark detectability and trajectory quality, we evaluate six $\delta$ values
ranging from 0 (no watermark) to 5 (maximum strength). We measure two key
metrics: Kullback-Leibler Divergence (KLD) quantifies how much the watermarked
distribution $\hat{D}_i$ deviates from the natural distribution $D_i$, while
TS assesses the quality of watermarked
action sequences using LLM-as-a-judge evaluation.
As shown in \T~\ref{tab:delta-impact}, increasing $\delta$ creates a trade-off
between watermark strength and trajectory quality. KLD grows exponentially from
0.096 ($\delta=1$) to 1.808 ($\delta=5$), indicating progressively stronger
distribution shifts. Concurrently, TS degrades from 0.712 to 0.671, revealing
that excessive bias introduces semantic drift in action sequences. 
This trend exposes a critical vulnerability: overly aggressive watermarking not only 
compromises functional quality but also increases statistical detectability,
potentially alerting sophisticated adversaries to the presence of embedded
signatures.
Based on these results, we recommend moderate $\delta$ values (2--3) for
practical deployment. This range achieves sufficient watermark detectability
while maintaining high trajectory fidelity, balancing
IP protection with service quality.
\section{Conclusion}
\label{sec:conclusion}

In this work, we uncover the susceptibility of commercial agentic systems to
imitation attacks and propose \tool, the first watermarking framework designed
for grey-box agentic models. By exploiting semantic equivalence in action
spaces, \tool injects imperceptible yet verifiable signatures into execution
trajectories without compromising service quality. Extensive evaluations across
diverse domains demonstrate that \tool achieves near-perfect detection and user
attribution accuracy while maintaining strong robustness against sophisticated
removal attempts. As agentic AI evolves, \tool provides a foundational and
practical safeguard for protecting proprietary model capabilities in the wild.

\bibliographystyle{IEEEtranN}

\bibliography{bib/references.bib}

@misc{openai2023,
  title={GPT-5 System Card},
  author={OpenAI},
  year={2025},
  url={https://openai.com/index/gpt-5-system-card/}
}

@article{xi2025rise,
  title={The rise and potential of large language model based agents: A survey},
  author={Xi, Zhiheng and Chen, Wenxiang and Guo, Xin and He, Wei and Ding, Yiwen and Hong, Boyang and Zhang, Ming and Wang, Junzhe and Jin, Senjie and Zhou, Enyu and others},
  journal={Science China Information Sciences},
  volume={68},
  number={2},
  pages={121101},
  year={2025},
  publisher={Springer}
}

@article{wang2024survey,
  title={A survey on large language model based autonomous agents},
  author={Wang, Lei and Ma, Chen and Feng, Xueyang and Zhang, Zeyu and Yang, Hao and Zhang, Jingsen and Chen, Zhiyuan and Tang, Jiakai and Chen, Xu and Lin, Yankai and others},
  journal={Frontiers of Computer Science},
  volume={18},
  number={6},
  pages={186345},
  year={2024},
  publisher={Springer}
}

@inproceedings{yao2022react,
  title={React: Synergizing reasoning and acting in language models},
  author={Yao, Shunyu and Zhao, Jeffrey and Yu, Dian and Du, Nan and Shafran, Izhak and Narasimhan, Karthik R and Cao, Yuan},
  booktitle={The eleventh international conference on learning representations},
  year={2022}
}

@article{patel2024datadreamer,
  title={Datadreamer: A tool for synthetic data generation and reproducible llm workflows},
  author={Patel, Ajay and Raffel, Colin and Callison-Burch, Chris},
  journal={arXiv preprint arXiv:2402.10379},
  year={2024}
}

@inproceedings{hong2025data,
  title={Data interpreter: An llm agent for data science},
  author={Hong, Sirui and Lin, Yizhang and Liu, Bang and Liu, Bangbang and Wu, Binhao and Zhang, Ceyao and Li, Danyang and Chen, Jiaqi and Zhang, Jiayi and Wang, Jinlin and others},
  booktitle={Findings of the Association for Computational Linguistics: ACL 2025},
  pages={19796--19821},
  year={2025}
}

@article{wang2024large,
  title={Large language models as urban residents: An llm agent framework for personal mobility generation},
  author={Wang, Jiawei and Jiang, Renhe and Yang, Chuang and Wu, Zengqing and Onizuka, Makoto and Shibasaki, Ryosuke and Koshizuka, Noboru and Xiao, Chuan},
  journal={Advances in Neural Information Processing Systems},
  volume={37},
  pages={124547--124574},
  year={2024}
}

@inproceedings{pan2025multiagent,
  title={Why do multiagent systems fail?},
  author={Pan, Melissa Z and Cemri, Mert and Agrawal, Lakshya A and Yang, Shuyi and Chopra, Bhavya and Tiwari, Rishabh and Keutzer, Kurt and Parameswaran, Aditya and Ramchandran, Kannan and Klein, Dan and others},
  booktitle={ICLR 2025 Workshop on Building Trust in Language Models and Applications},
  year={2025}
}

@misc{openai2025o3,
  title={OpenAI o3 system card},
  author={OpenAI},
  year={2025},
  url={https://openai.com/index/introducing-o3-and-o4-mini/}
}

@article{guo2025deepseek,
  title={Deepseek-r1: Incentivizing reasoning capability in llms via reinforcement learning},
  author={Guo, Daya and Yang, Dejian and Zhang, Haowei and Song, Junxiao and Zhang, Ruoyu and Xu, Runxin and Zhu, Qihao and Ma, Shirong and Wang, Peiyi and Bi, Xiao and others},
  journal={arXiv preprint arXiv:2501.12948},
  year={2025}
}

@misc{minimax2025,
  title={Minimax 2.1 system card},
  author={Minimax},
  year={2025},
  url={https://www.minimax.io/news/minimax-m21}
}

@inproceedings{zhao2025survey,
  title={A survey on model extraction attacks and defenses for large language models},
  author={Zhao, Kaixiang and Li, Lincan and Ding, Kaize and Gong, Neil Zhenqiang and Zhao, Yue and Dong, Yushun},
  booktitle={Proceedings of the 31st ACM SIGKDD Conference on Knowledge Discovery and Data Mining V. 2},
  pages={6227--6236},
  year={2025}
}

@inproceedings{liu2025evaluating,
  title={Evaluating $\{$LLM-based$\}$ Personal Information Extraction and Countermeasures},
  author={Liu, Yupei and Jia, Yuqi and Jia, Jinyuan and Gong, Neil Zhenqiang},
  booktitle={34th USENIX Security Symposium (USENIX Security 25)},
  pages={1669--1688},
  year={2025}
}

@inproceedings{carlini2024stealing,
  title={Stealing part of a production language model},
  author={Carlini, Nicholas and Paleka, Daniel and Dvijotham, Krishnamurthy and Steinke, Thomas and Hayase, Jonathan and Cooper, A Feder and Lee, Katherine and Jagielski, Matthew and Nasr, Milad and Conmy, Arthur and others},
  booktitle={Proceedings of the 41st International Conference on Machine Learning},
  pages={5680--5705},
  year={2024}
}

@inproceedings{li2025differentiation,
  title={Differentiation-based extraction of proprietary data from fine-tuned llms},
  author={Li, Zongjie and Wu, Daoyuan and Wang, Shuai and Su, Zhendong},
  booktitle={Proceedings of the 2025 ACM SIGSAC Conference on Computer and Communications Security},
  pages={3071--3085},
  year={2025}
}

@inproceedings{li2023protecting,
  title={Protecting intellectual property of large language model-based code generation apis via watermarks},
  author={Li, Zongjie and Wang, Chaozheng and Wang, Shuai and Gao, Cuiyun},
  booktitle={Proceedings of the 2023 ACM SIGSAC Conference on Computer and Communications Security},
  pages={2336--2350},
  year={2023}
}

@article{he2022cater,
  title={Cater: Intellectual property protection on text generation apis via conditional watermarks},
  author={He, Xuanli and Xu, Qiongkai and Zeng, Yi and Lyu, Lingjuan and Wu, Fangzhao and Li, Jiwei and Jia, Ruoxi},
  journal={Advances in Neural Information Processing Systems},
  volume={35},
  pages={5431--5445},
  year={2022}
}

@inproceedings{zhang2024remark,
  title={$\{$REMARK-LLM$\}$: A robust and efficient watermarking framework for generative large language models},
  author={Zhang, Ruisi and Hussain, Shehzeen Samarah and Neekhara, Paarth and Koushanfar, Farinaz},
  booktitle={33rd USENIX Security Symposium (USENIX Security 24)},
  pages={1813--1830},
  year={2024}
}

@inproceedings{zhao2025can,
  title={Can watermarks be used to detect LLM IP infringement for free?},
  author={Zhao, Zhengyue and Liu, Xiaogeng and Jha, Somesh and McDaniel, Patrick and Li, Bo and Xiao, Chaowei},
  booktitle={The Thirteenth International Conference on Learning Representations},
  year={2025}
}

@article{Dathathri2024,
    author={Dathathri, Sumanth and See, Abigail and Ghaisas, Sumedh and Huang, Po-Sen and McAdam, Rob and Welbl, Johannes and Bachani, Vandana and Kaskasoli, Alex and Stanforth, Robert and Matejovicova, Tatiana and Hayes, Jamie and Vyas, Nidhi and Merey, Majd Al and Brown-Cohen, Jonah and Bunel, Rudy and Balle, Borja and Cemgil, Taylan and Ahmed, Zahra and Stacpoole, Kitty and Shumailov, Ilia and Baetu, Ciprian and Gowal, Sven and Hassabis, Demis and Kohli, Pushmeet},
    title={Scalable watermarking for identifying large language model outputs},
    journal={Nature},
    url={https://doi.org/10.1038/s41586-024-08025-4}
}

@article{alon2023detecting,
  title={Detecting language model attacks with perplexity},
  author={Alon, Gabriel and Kamfonas, Michael},
  journal={arXiv preprint arXiv:2308.14132},
  year={2023}
}

@article{brown2020language,
  title={Language models are few-shot learners},
  author={Brown, Tom and Mann, Benjamin and Ryder, Nick and Subbiah, Melanie and Kaplan, Jared D and Dhariwal, Prafulla and Neelakantan, Arvind and Shyam, Pranav and Sastry, Girish and Askell, Amanda and others},
  journal={Advances in neural information processing systems},
  volume={33},
  pages={1877--1901},
  year={2020}
}

@article{chung2024scaling,
  title={Scaling instruction-finetuned language models},
  author={Chung, Hyung Won and Hou, Le and Longpre, Shayne and Zoph, Barret and Tay, Yi and Fedus, William and Li, Yunxuan and Wang, Xuezhi and Dehghani, Mostafa and Brahma, Siddhartha and others},
  journal={Journal of Machine Learning Research},
  volume={25},
  number={70},
  pages={1--53},
  year={2024}
}

@article{ouyang2022training,
  title={Training language models to follow instructions with human feedback},
  author={Ouyang, Long and Wu, Jeffrey and Jiang, Xu and Almeida, Diogo and Wainwright, Carroll and Mishkin, Pamela and Zhang, Chong and Agarwal, Sandhini and Slama, Katarina and Ray, Alex and others},
  journal={Advances in neural information processing systems},
  volume={35},
  pages={27730--27744},
  year={2022}
}

@article{schulman2017proximal,
  title={Proximal policy optimization algorithms},
  author={Schulman, John and Wolski, Filip and Dhariwal, Prafulla and Radford, Alec and Klimov, Oleg},
  journal={arXiv preprint arXiv:1707.06347},
  year={2017}
}

@article{zhou2023lima,
  title={Lima: Less is more for alignment},
  author={Zhou, Chunting and Liu, Pengfei and Xu, Puxin and Iyer, Srinivasan and Sun, Jiao and Mao, Yuning and Ma, Xuezhe and Efrat, Avia and Yu, Ping and Yu, Lili and others},
  journal={Advances in Neural Information Processing Systems},
  volume={36},
  pages={55006--55021},
  year={2023}
}

@article{ye2025limo,
  title={Limo: Less is more for reasoning},
  author={Ye, Yixin and Huang, Zhen and Xiao, Yang and Chern, Ethan and Xia, Shijie and Liu, Pengfei},
  journal={arXiv preprint arXiv:2502.03387},
  year={2025}
}

@misc{coze2025,
  title={Coze platform},
  url={https://www.coze.com/}
}

@misc{dify2025,
  title={Dify platform},
  url={https://dify.ai//}
}

@article{wang2025ip,
  title={Ip leakage attacks targeting llm-based multi-agent systems},
  author={Wang, Liwen and Wang, Wenxuan and Wang, Shuai and Li, Zongjie and Ji, Zhenlan and Lyu, Zongyi and Wu, Daoyuan and Cheung, Shing-Chi},
  journal={arXiv preprint arXiv:2505.12442},
  year={2025}
}

@article{li2025dissonances,
  title={Les Dissonances: Cross-Tool Harvesting and Polluting in Multi-Tool Empowered LLM Agents},
  author={Li, Zichuan and Cui, Jian and Liao, Xiaojing and Xing, Luyi},
  journal={NDSS 2026}
}

@inproceedings{li2026webcloak,
  title={WebCloak: Characterizing and Mitigating the Threats of LLM-Driven Web Agents as Intelligent Scrapers},
  author={Li, Xinfeng and Qiu, Tianze and Jin, Yingbin and Wang, Lixu and Guo, Hanqing and Jia, Xiaojun and Wang, XiaoFeng and Dong, Wei},
  booktitle={IEEE Symposium on Security and Privacy},
  year={2026}
}

@inproceedings{carlini2023extracting,
  title={Extracting training data from diffusion models},
  author={Carlini, Nicolas and Hayes, Jamie and Nasr, Milad and Jagielski, Matthew and Sehwag, Vikash and Tramer, Florian and Balle, Borja and Ippolito, Daphne and Wallace, Eric},
  booktitle={32nd USENIX security symposium (USENIX Security 23)},
  pages={5253--5270},
  year={2023}
}

@article{yang2025qwen3,
  title={Qwen3 technical report},
  author={Yang, An and Li, Anfeng and Yang, Baosong and Zhang, Beichen and Hui, Binyuan and Zheng, Bo and Yu, Bowen and Gao, Chang and Huang, Chengen and Lv, Chenxu and others},
  journal={arXiv preprint arXiv:2505.09388},
  year={2025}
}

@inproceedings{chandrasekaran2020exploring,
  title={Exploring connections between active learning and model extraction},
  author={Chandrasekaran, Varun and Chaudhuri, Kamalika and Giacomelli, Irene and Jha, Somesh and Yan, Songbai},
  booktitle={29th USENIX Security Symposium (USENIX Security 20)},
  pages={1309--1326},
  year={2020}
}

@misc{manus2025,
  title={Manus platform},
  url={https://www.manus.ai/}
}

@article{liu2024apigen,
  title={Apigen: Automated pipeline for generating verifiable and diverse function-calling datasets},
  author={Liu, Zuxin and Hoang, Thai and Zhang, Jianguo and Zhu, Ming and Lan, Tian and Tan, Juntao and Yao, Weiran and Liu, Zhiwei and Feng, Yihao and RN, Rithesh and others},
  journal={Advances in Neural Information Processing Systems},
  volume={37},
  pages={54463--54482},
  year={2024}
}

@article{liu2025deepseek,
  title={Deepseek-v3. 2: Pushing the frontier of open large language models},
  author={Liu, Aixin and Mei, Aoxue and Lin, Bangcai and Xue, Bing and Wang, Bingxuan and Xu, Bingzheng and Wu, Bochao and Zhang, Bowei and Lin, Chaofan and Dong, Chen and others},
  journal={arXiv preprint arXiv:2512.02556},
  year={2025}
}

@article{yao2024tau,
  title={{$\tau$-bench: A Benchmark for Tool-Agent-User Interaction in Real-World Domains}},
  author={Yao, Shunyu and Shinn, Noah and Razavi, Pedram and Narasimhan, Karthik},
  journal={ICLR 2025},
  year={2024}
}

@inproceedings{qintoolllm,
  title={ToolLLM: Facilitating Large Language Models to Master 16000+ Real-world APIs},
  author={Qin, Yujia and Liang, Shihao and Ye, Yining and Zhu, Kunlun and Yan, Lan and Lu, Yaxi and Lin, Yankai and Cong, Xin and Tang, Xiangru and Qian, Bill and others},
  booktitle={The Twelfth International Conference on Learning Representations}
}

@article{chang2024agentboard,
  title={Agentboard: An analytical evaluation board of multi-turn llm agents},
  author={Chang, Ma and Zhang, Junlei and Zhu, Zhihao and Yang, Cheng and Yang, Yujiu and Jin, Yaohui and Lan, Zhenzhong and Kong, Lingpeng and He, Junxian},
  journal={Advances in neural information processing systems},
  volume={37},
  pages={74325--74362},
  year={2024}
}

@inproceedings{zhugeagent,
  title={Agent-as-a-Judge: Evaluate Agents with Agents},
  author={Zhuge, Mingchen and Zhao, Changsheng and Ashley, Dylan R and Wang, Wenyi and Khizbullin, Dmitrii and Xiong, Yunyang and Liu, Zechun and Chang, Ernie and Krishnamoorthi, Raghuraman and Tian, Yuandong and others},
  booktitle={ICML 2024}
}

@article{clark2018think,
  title={Think you have solved question answering? try arc, the ai2 reasoning challenge},
  author={Clark, Peter and Cowhey, Isaac and Etzioni, Oren and Khot, Tushar and Sabharwal, Ashish and Schoenick, Carissa and Tafjord, Oyvind},
  journal={arXiv preprint arXiv:1803.05457},
  year={2018}
}

@inproceedings{kokel2025acpbench,
  title={Acpbench: Reasoning about action, change, and planning},
  author={Kokel, Harsha and Katz, Michael and Srinivas, Kavitha and Sohrabi, Shirin},
  booktitle={Proceedings of the AAAI Conference on Artificial Intelligence},
  volume={39},
  number={25},
  pages={26559--26568},
  year={2025}
}

@article{kang2025distilling,
  title={Distilling llm agent into small models with retrieval and code tools},
  author={Kang, Minki and Jeong, Jongwon and Lee, Seanie and Cho, Jaewoong and Hwang, Sung Ju},
  journal={arXiv preprint arXiv:2505.17612},
  year={2025}
}

@misc{qwen3next2025,
  title={Qwen3-Next-80B-A3B},
  url={https://huggingface.co/collections/Qwen/qwen3-next}
}

@article{hu2022lora,
  title={Lora: Low-rank adaptation of large language models.},
  author={Hu, Edward J and Shen, Yelong and Wallis, Phillip and Allen-Zhu, Zeyuan and Li, Yuanzhi and Wang, Shean and Wang, Lu and Chen, Weizhu and others},
  journal={ICLR},
  volume={1},
  number={2},
  pages={3},
  year={2022}
}

@inproceedings{chenmark,
  title={De-mark: Watermark Removal in Large Language Models},
  author={Chen, Ruibo and Wu, Yihan and Guo, Junfeng and Huang, Heng},
  booktitle={Forty-second International Conference on Machine Learning}
}

@inproceedings{wangtowards,
  title={Towards Codable Watermarking for Injecting Multi-Bits Information to LLMs},
  author={Wang, Lean and Yang, Wenkai and Chen, Deli and Zhou, Hao and Lin, Yankai and Meng, Fandong and Zhou, Jie and Sun, Xu},
  booktitle={The Twelfth International Conference on Learning Representations}
}

@inproceedings{kirchenbauer2023watermark,
  title={A watermark for large language models},
  author={Kirchenbauer, John and Geiping, Jonas and Wen, Yuxin and Katz, Jonathan and Miers, Ian and Goldstein, Tom},
  booktitle={International Conference on Machine Learning},
  pages={17061--17084},
  year={2023},
  organization={PMLR}
}

@inproceedings{liu2024adaptive,
  title={Adaptive text watermark for large language models},
  author={Liu, Yepeng and Bu, Yuheng},
  booktitle={Proceedings of the 41st International Conference on Machine Learning},
  pages={30718--30737},
  year={2024}
}

@inproceedings{liusemantic,
  title={A Semantic Invariant Robust Watermark for Large Language Models},
  author={Liu, Aiwei and Pan, Leyi and Hu, Xuming and Meng, Shiao and Wen, Lijie},
  booktitle={The Twelfth International Conference on Learning Representations}
}

@inproceedings{hou2024semstamp,
  title={Semstamp: A semantic watermark with paraphrastic robustness for text generation},
  author={Hou, Abe and Zhang, Jingyu and He, Tianxing and Wang, Yichen and Chuang, Yung-Sung and Wang, Hongwei and Shen, Lingfeng and Van Durme, Benjamin and Khashabi, Daniel and Tsvetkov, Yulia},
  booktitle={NAACL},
  pages={4067--4082},
  year={2024}
}

@inproceedings{fu2024gumbelsoft,
  title={GumbelSoft: Diversified Language Model Watermarking via the GumbelMax-trick},
  author={Fu, Jiayi and Zhao, Xuandong and Yang, Ruihan and Zhang, Yuansen and Chen, Jiangjie and Xiao, Yanghua},
  booktitle={ACL},
}

@article{wen2025cotguard,
  title={CoTGuard: Using Chain-of-Thought Triggering for Copyright Protection in Multi-Agent LLM Systems},
  author={Wen, Yan and Guo, Junfeng and Huang, Heng},
  journal={arXiv preprint arXiv:2505.19405},
  year={2025}
}

@article{huang2025agent,
  title={Agent guide: A simple agent behavioral watermarking framework},
  author={Huang, Kaibo and Zhang, Zipei and Yang, Zhongliang and Zhou, Linna},
  journal={arXiv preprint arXiv:2504.05871},
  year={2025}
}

@article{yoon2025intrinsic,
  title={Intrinsic Fingerprint of LLMs: Continue Training is NOT All You Need to Steal A Model!},
  author={Yoon, Do-hyeon and Chun, Minsoo and Allen, Thomas and M{\"u}ller, Hans and Wang, Min and Sharma, Rajesh},
  journal={arXiv preprint arXiv:2507.03014},
  year={2025}
}

@article{shao2024explanation,
  title={Explanation as a watermark: Towards harmless and multi-bit model ownership verification via watermarking feature attribution},
  author={Shao, Shuo and Li, Yiming and Yao, Hongwei and He, Yiling and Qin, Zhan and Ren, Kui},
  journal={arXiv preprint arXiv:2405.04825},
  year={2024}
}

@inproceedings{xu2024instructional,
  title={Instructional fingerprinting of large language models},
  author={Xu, Jiashu and Wang, Fei and Ma, Mingyu and Koh, Pang Wei and Xiao, Chaowei and Chen, Muhao},
  booktitle={Proceedings of the 2024 Conference of the North American Chapter of the Association for Computational Linguistics: Human Language Technologies (Volume 1: Long Papers)},
  pages={3277--3306},
  year={2024}
}

@misc{mistral2024,
  title={Ministral-8B-Instruct-2410},
  url={https://huggingface.co/mistralai/Ministral-8B-Instruct-2410}
}

@inproceedings{gilbert2003security,
  title={Security analysis of SHA-256 and sisters},
  author={Gilbert, Henri and Handschuh, Helena},
  booktitle={International workshop on selected areas in cryptography},
  pages={175--193},
  year={2003},
  organization={Springer}
}

@misc{qwenembedding2025,
  title={Qwen3-Embedding-0.6B},
  url={https://huggingface.co/Qwen/Qwen3-Embedding-0.6B}
}

@article{ji2025tree,
  title={Tree search for llm agent reinforcement learning},
  author={Ji, Yuxiang and Ma, Ziyu and Wang, Yong and Chen, Guanhua and Chu, Xiangxiang and Wu, Liaoni},
  journal={arXiv preprint arXiv:2509.21240},
  year={2025}
}

@inproceedings{chen2025optima,
  title={Optima: Optimizing effectiveness and efficiency for llm-based multi-agent system},
  author={Chen, Weize and Yuan, Jiarui and Qian, Chen and Yang, Cheng and Liu, Zhiyuan and Sun, Maosong},
  booktitle={Findings of the Association for Computational Linguistics: ACL 2025},
  pages={11534--11557},
  year={2025}
}

@inproceedings{zhai2025enhancing,
  title={Enhancing decision-making for llm agents via step-level q-value models},
  author={Zhai, Yuanzhao and Yang, Tingkai and Xu, Kele and Feng, Dawei and Yang, Cheng and Ding, Bo and Wang, Huaimin},
  booktitle={Proceedings of the AAAI Conference on Artificial Intelligence},
}

@article{koh2024tree,
  title={Tree search for language model agents},
  author={Koh, Jing Yu and McAleer, Stephen and Fried, Daniel and Salakhutdinov, Ruslan},
  journal={arXiv preprint arXiv:2407.01476},
  year={2024}
}

@article{radford2019language,
  title={Language models are unsupervised multitask learners},
  author={Radford, Alec and Wu, Jeffrey and Child, Rewon and Luan, David and Amodei, Dario and Sutskever, Ilya and others},
  journal={OpenAI blog},
  volume={1},
  number={8},
  pages={9},
  year={2019}
}

@article{shen2025efficient,
  title={Efficient reasoning with hidden thinking},
  author={Shen, Xuan and Wang, Yizhou and Shi, Xiangxi and Wang, Yanzhi and Zhao, Pu and Gu, Jiuxiang},
  journal={arXiv preprint arXiv:2501.19201},
  year={2025}
}

@inproceedings{yang2025prsa,
  title={{PRSA}: Prompt stealing attacks against {Real-World} prompt services},
  author={Yang, Yong and Li, Changjiang and Li, Qingming and Ma, Oubo and Wang, Haoyu and Wang, Zonghui and Gao, Yandong and Chen, Wenzhi and Ji, Shouling},
  booktitle={34th USENIX security symposium (USENIX Security 25)},
  pages={2283--2302},
  year={2025}
}

@inproceedings {carlini2021extracting,
author = {Nicholas Carlini and Florian Tram{\`e}r and Eric Wallace and Matthew Jagielski and Ariel Herbert-Voss and Katherine Lee and Adam Roberts and Tom Brown and Dawn Song and {\'U}lfar Erlingsson and Alina Oprea and Colin Raffel},
title = {Extracting Training Data from Large Language Models},
booktitle = {30th USENIX Security Symposium (USENIX Security 21)},
}

@article{zou2023universal,
  title={Universal and transferable adversarial attacks on aligned language models},
  author={Zou, Andy and Wang, Zifan and Carlini, Nicholas and Nasr, Milad and Kolter, J Zico and Fredrikson, Matt},
  journal={arXiv preprint arXiv:2307.15043},
  year={2023}
}

@article{yi2024jailbreak,
  title={Jailbreak attacks and defenses against large language models: A survey},
  author={Yi, Sibo and Liu, Yule and Sun, Zhen and Cong, Tianshuo and He, Xinlei and Song, Jiaxing and Xu, Ke and Li, Qi},
  journal={arXiv preprint arXiv:2407.04295},
  year={2024}
}

@article{goodfellow2018making,
  title={Making machine learning robust against adversarial inputs},
  author={Goodfellow, Ian and McDaniel, Patrick and Papernot, Nicolas},
  journal={Communications of the ACM},
  volume={61},
  number={7},
  pages={56--66},
  year={2018},
  publisher={ACM New York, NY, USA}
}

@article{goodfellow2014explaining,
  title={Explaining and harnessing adversarial examples},
  author={Goodfellow, Ian J and Shlens, Jonathon and Szegedy, Christian},
  journal={arXiv preprint arXiv:1412.6572},
  year={2014}
}

@inproceedings{hendrycksmeasuring,
  title={Measuring Massive Multitask Language Understanding},
  author={Hendrycks, Dan and Burns, Collin and Basart, Steven and Zou, Andy and Mazeika, Mantas and Song, Dawn and Steinhardt, Jacob},
  booktitle={International Conference on Learning Representations}
}

@inproceedings{valmeekam2022large,
  title={Large language models still can't plan (a benchmark for LLMs on planning and reasoning about change)},
  author={Valmeekam, Karthik and Olmo, Alberto and Sreedharan, Sarath and Kambhampati, Subbarao},
  booktitle={NeurIPS 2022 Foundation Models for Decision Making Workshop},
  year={2022}
}

\appendix
\raggedbottom

\section{Platform Visibility Survey}
\label{appendix:platform-survey}

To empirically validate the prevalence of grey-box visibility constraints in
commercial agentic systems, we conducted a comprehensive survey of 29 mainstream
agentic platforms and frameworks. We categorize each platform based on its
default observability characteristics into three visibility levels:

\begin{itemize}[leftmargin=*,itemsep=2pt]
    \item \textbf{White-box}: Systems that expose complete, token-level trajectory
    of LLM reasoning, including full input/output logs, intermediate
    chain-of-thought, and fine-grained execution details. These platforms
    typically support streaming tokens and comprehensive tracing infrastructure.
    
    \item \textbf{Grey-box}: Systems that reveal action sequences and summarized
    reasoning steps for billing and debugging purposes, but conceal raw
    token-level internal reasoning traces. This represents the dominant
    deployment model for commercial agentic services.
    
    \item \textbf{Black-box}: Systems that only expose final outputs with
    minimal visibility into intermediate execution steps or tool invocations.
\end{itemize}

\T~\ref{tab:platform-survey} presents our classification results. We observe
that 24 out of 29 platforms (82.8\%) adopt the grey-box model, confirming our
threat model assumption in \S~\ref{subsec:threat-model}. This visibility
constraint fundamentally motivates \tool's design: existing watermarking
techniques that rely on embedding signals into complete reasoning traces fail
under grey-box constraints, necessitating our action-sequence-based approach.

\begin{table}[!t]
    \centering
    \caption{Classification of 29 mainstream agentic platforms by visibility level. Platforms are categorized based on their default observability characteristics without additional third-party instrumentation.}
    \label{tab:platform-survey}
    \resizebox{0.95\linewidth}{!}{
    \begin{tabular}{lcp{8cm}}
    \toprule
    \textbf{Visibility} & \textbf{Count} & \textbf{Platforms Name} \\
    \midrule
    \textbf{White-box} & 3 & 
    LangGraph, Langfuse, PydanticAI\\
    \midrule
    \textbf{Grey-box} & 24 & Coze, Dify.ai, CrewAI, Semantic Kernel, Microsoft
    Copilot Studio, Salesforce Agentforce, Google Vertex AI, Relevance AI, Dust,
    Lyzr, Cognosys, Glean Agents, Taskade, Make.com, Vellum AI, Genspark Super
    Agent, Manus AI, LangFlow, Boomi AI Studio, Sierra, n8n, Fiddler AI,
    Microsoft Agent Framework, Google Gemini \\
    \midrule
    \textbf{Black-box} & 2 & Devin, Workday Illuminate
    \\
    \bottomrule
    \end{tabular}
    }
\end{table}

\end{document}